\documentclass[11pt]{article} % use larger type; default would be 10pt

\usepackage[utf8]{inputenc} % set input encoding (not needed with XeLaTeX)
\usepackage{amsthm,amsmath}
\usepackage{geometry} % to change the page dimensions
\usepackage{graphicx} % support the \includegraphics command and options
\usepackage{url} 
\usepackage{hyperref}
\usepackage{authblk}
\usepackage[english]{babel}
\usepackage{booktabs} % for much better looking tables
\usepackage{array} % for better arrays (eg matrices) in maths
\usepackage{paralist} % very flexible & customisable lists (eg. enumerate/itemize, etc.)
\usepackage{verbatim} % adds environment for commenting out blocks of text & for better verbatim
\usepackage{subfig} % make it possible to include more than one captioned figure/table in a single float
\usepackage{natbib}
\usepackage{fancyhdr} % This should be set AFTER setting up the page geometry
\usepackage{sectsty}
\allsectionsfont{\sffamily\mdseries\upshape} % (See the fntguide.pdf for font help)
\usepackage[nottoc,notlof,notlot]{tocbibind} % Put the bibliography in the ToC
\usepackage[titles,subfigure]{tocloft} % Alter the style of the Table of Contents

\title{A Systematic Review of Natural Language Processing Applied to Radiology Reports}

\author[1,*]{Arlene Casey}
\author[2]{Emma Davidson}
\author[2]{Michael Poon}
\author[3,4]{Hang Dong}
\author[1]{Daniel Duma}
\author[5]{Andreas Grivas}
\author[5]{Claire Grover}
\author[3,4]{Víctor Suárez-Paniagua}
\author[5]{Richard Tobin}
\author[2,6]{William Whiteley}
\author[4,7]{Honghan Wu}
\author[1,8]{Beatrice Alex}
\affil[1]{School of Literatures, Languages and Cultures (LLC), University of Edinburgh} 
\affil[2]{Centre for Clinical Brain Sciences, University of Edinburgh} 
\affil[3]{Centre for Medical Informatics, Usher Institute of Population Health Sciences and Informatics, University of Edinburgh}
\affil[4]{Health Data Research, UK}
\affil[5]{Institute for Language, Cognition and Computation, School of Informatics, University of Edinburgh}
\affil[6]{Nuffield Department of Population Health, University of Oxford}
\affil[7]{Institute of Health Informatics, University College of London}
\affil[8]{Edinburgh Futures Institute, University of Edinburgh}
\affil[*]{Corresponding author:arlene.casey AT ed.ac.uk}

\date{\vspace{-5ex}}

\begin{document}
\maketitle

%\begin{abstractbox}
\begin{abstract} % abstract
\textbf{Background} %if any
Natural language processing (NLP) has a significant role in advancing healthcare and has been found to be key in extracting structured information from radiology reports.  Understanding recent developments in NLP application to radiology is of significance but recent reviews on this are limited.  This study systematically assesses and quantifies recent literature in NLP applied to radiology reports. 

\textbf{Methods} We conduct an automated literature search yielding 4,799 results using automated filtering, metadata enriching steps and citation search combined with manual review. Our analysis is based on 21 variables including radiology characteristics, NLP methodology, performance, study, and clinical application characteristics.

\textbf{Results} We present a comprehensive analysis of the 164 publications retrieved with publications in 2019 almost triple those in 2015. Each publication is categorised into one of 6 clinical application categories. Deep learning use increases in the period but conventional machine learning approaches are still prevalent. Deep learning remains challenged when data is scarce and there is little evidence of adoption into clinical practice. Despite 17\% of studies reporting greater than 0.85 F1 scores, it is hard to comparatively evaluate these approaches given that most of them use different datasets. Only 14 studies made their data and 15 their code available with 10 externally validating results.      

\textbf{Conclusions}  Automated understanding of clinical narratives of the radiology reports has the potential to enhance the healthcare process and we show that research in this field continues to grow. Reproducibility and explainability of models are important if the domain is to move applications into clinical use. More could be done to share code enabling validation of methods on different institutional data and to reduce heterogeneity in reporting of study properties allowing inter-study comparisons.  Our results have significance for researchers in the field providing a systematic synthesis of existing work to build on, identify gaps, opportunities for collaboration and avoid duplication. 
\end{abstract}

\section*{Background}

Medical imaging examinations interpreted by radiologists in the form of narrative reports are used to support and confirm diagnosis in clinical practice. Being able to accurately and quickly identify the information stored in radiologists' narratives has the potential to reduce workloads, support clinicians in their decision processes, triage patients to get urgent care or identify patients for research purposes. However, whilst these reports are generally considered more restricted in vocabulary than other electronic health records (EHR),  e.g. clinical notes, it is still difficult to access this efficiently at scale \citep{bates_classification_2016}. This is due to the unstructured nature of these reports and Natural Language Processing (NLP) is key to obtaining structured information from radiology reports \citep{pons_natural_2016}.

NLP applied to radiology reports is shown to be a growing field in earlier reviews \citep{pons_natural_2016,cai_natural_2016}.  
In recent years there has been an even more extensive growth in NLP research in general and in particular deep learning methods which is not seen in the earlier reviews.  A more recent review of NLP applied to radiology-related research can be found but this focuses on one NLP technique only, deep learning models \citep{sorin_deep_2020}. Our paper provides a more comprehensive review comparing and contrasting all NLP methodologies as they are applied to radiology.  

It is of significance to understand and synthesise recent developments specific to NLP in the radiology research field as this will assist researchers to gain a broader understanding of the field, provide insight into methods and techniques supporting and promoting new developments in the field. Therefore, we carry out a systematic review of research output on NLP applications in radiology from 2015 onward, thus, allowing for a more up to date analysis of the area. 
%edited this for arxiv
%We include as an additional file our synthesis of the publications detailing their clinical and technical categories along with anatomical scan regions.
An additional listing of our synthesis of publications detailing their clinical and technical categories along with anatomical scan regions can be made available on request.
Also different to the existing work, we look at both the clinical application areas NLP is being applied in and consider the trends in NLP methods.  We describe and discuss study properties, e.g. data size, performance, annotation details, quantifying these in relation to both the clinical application areas and NLP methods.  Having a more detailed understanding of these properties allows us to make recommendations for future NLP research applied to radiology datasets, supporting improvements and progress in this domain.

%%%%%%%%%%%%%%%%%%%%%%%%%%%%%%%%%%%%%%%%%%%%%%%%%%%%%%%%%%%%%%%%%%%%%%%%%%%%%%%%%%%%%%%%%%%%%%%%%%%%%%%%%%%%%%
\section*{Related Work}
Amongst pre-existing reviews in this area, \citep{pons_natural_2016} was the first that was both specific to NLP on radiology reports and systematic in methodology. Their literature search identified 67 studies published in the period up to October 2014. They examined the NLP methods used, summarised their performance and extracted the studies' clinical applications, which they assigned to five broad categories delineating their purpose. Since Pons et al.'s paper, several reviews have emerged with the broader remit of NLP applied to electronic health data, which includes radiology reports. \citep{kreimeyer_natural_2017} conducted a systematic review of NLP systems with a specific focus on coding free text into clinical terminologies and structured data capture. The systematic review by \citep{spasic_clinical_2020} specifically examined machine learning approaches to NLP (2015-2019) in more general clinical text data, and a further methodical review was carried out by \citep{wu_deep_2020} to synthesise literature on deep learning in clinical NLP (up to April 2019) although the did not follow the PRISMA guideline completely. With radiology reports as their particular focus, \citep{cai_natural_2016} published, the same year as Pons et al.'s review, an instructive narrative review outlining the fundamentals of NLP techniques applied in radiology. More recently, \citep{sorin_deep_2020} published a systematic review focused on deep learning radiology-related research. They identified 10 relevant papers in their search (up to September 2019) and examined their deep learning models, comparing these with traditional NLP models and also considered their clinical applications but did not employ a specific categorisation. We build on this corpus of related work, and most specifically Pons et al.'s work. In our initial synthesis of clinical applications we adopt their application categories and further expand upon these to reflect the nature of subsequent literature captured in our work.  Additionally, we quantify and compare properties of the studies reviewed and provide a series of recommendations for future NLP research applied to radiology datasets in order to promote improvements and progress in this domain.

\section*{Methods}

Our methodology followed the Preferred Reporting Items for Systematic Reviews and Meta-Analysis (PRISMA) \citep{moher_preferred_2015}, and the protocol is registered on protocols.io. 
\subsection*{Eligibility for Literature Inclusion and Search Strategy}
We included studies using NLP on radiology reports of any imaging modality and anatomical region for NLP technical development, clinical support, or epidemiological research. Exclusion criteria included: (1) case reports; (2) published before 2015; (3) in language other than English; (4) processing of radiology images; (5) reviews, conference abstracts, comments, patents, or editorials; (6) not reporting outcomes of interest; (7) not radiology reports; (8) not using NLP methods; (9) not available in full text; (10) duplicates.

We used Publish or Perish \citep{harzing_a_w_publish_2007}, a citation retrieval and analysis software program,  to search Google Scholar. Google Scholar has a similar coverage to other databases \citep{gehanno_is_2013} and is easier to integrate into search pipelines. We conducted an initial pilot search following the process described here, but the search terms were too specific and restricted the number of publications.  However, we did include papers found in the pilot search in full-text screening. 
We use the following search query restricted to research articles published in English between January 2015 and October 2019.  
("radiology" OR "radiologist") AND ("natural language" OR "text mining" OR "information extraction" OR "document classification" OR "word2vec") NOT patent. 
We automated the addition of publication metadata and applied filtering to remove irrelevant publications.  These automated steps are described in Table \ref{tab:t1} \& Table \ref{tab:t2}.
%TABLE 1
\begin{table}[h!]
\small
\caption{Metadata enriching steps undertaken for each publication}
  \begin{tabular}{l}
  
  Metadata Enriching Steps \\
    \hline
1.	Match the paper with its DOI via the Crossref API%\citep{wilkinson_rest_nodate}
 \\
2.	If  DOI matched, check Semantic Scholar  for metadata/abstract %\citep{noauthor_scholar_nodate}
 \\
3.	If no DOI match and no abstract, search PubMed for abstract\\
4.	Search arXiv% \citep{noauthor_arxivorg_nodate} 
(for a pre-print)\\
5.	If no PDF link, search Unpaywall  for available open access versions %\citep{bearden_libguides_nodate}
  \\
6.	If PDF but no separate abstract via Semantics Scholar/PubMed, extract abstract from the PDF\\
  \end{tabular}
  \label{tab:t1}
\end{table}
%Table 2
\begin{table}[h!]
\small
\caption{Automated filtering steps to remove irrelevant publications}
  \begin{tabular}{l}
  Automated Filtering Steps\\
    \hline
1.	Document language is English\\
2.	Word 'patent' in title or URL\\
3.	Year of publication out of range (\textless 2015)\\
4.	The words 'review' or 'overview' in the title, 'this review' in the abstract\\
5.	Image keywords in title or abstract \\
6.	No radiology keywords in title or abstract \\
7.	No NLP terminology in abstract\\
  \end{tabular}
  \label{tab:t2}
\end{table}

In addition to query search, another method to find papers is to conduct a citation search \citep{briscoe_conduct_2020}. The citation search compiled a list of publications that cite the Pons et al. review and the articles cited in the Pons’ review.  To do this, we use a snowballing method \citep{wohlin_guidelines_2014} to follow the forward citation branch for each publication in this list, i.e. finding every article that cites the publications in our list.  The branching factor here is large, so we filter at every stage and automatically add metadata.
\subsection*{Manual Review of Literature}
Four reviewers (three NLP researchers [AG,DD and HD] and one epidemiologist [MTCP]) independently screened all titles and abstracts with the Rayyan  online platform and discussed disagreements. Fleiss' kappa \citep{fleiss_measuring_1971} agreement between reviewers was 0.70, indicating substantial agreement \citep{landis_measurement_1977}.  After this screening process, each full-text article was reviewed by a team of eight (six NLP researchers and two epidemiologists) and double reviewed by a NLP researcher. We resolved any discrepancies by discussion in regular meetings.

\subsection*{Data Extraction for Analysis}
We extracted data on: primary clinical application and technical objective, data source(s), study period, radiology report language, anatomical region, imaging modality, disease area, dataset size, annotated set size, training/validation/test set size, external validation performed, domain expert used, number of annotators, inter-annotator agreement, NLP technique(s) used, best-reported results (recall, precision and F1 score), availability of dataset, and availability of code. 

%In recording NLP technique we capture whether the author uses rules, machine learning methods, deep learning, ontologies, lexicons and word embeddings.  We discriminate machine learning from deep learning, using the former to represent traditional machine learning methods.

\section*{Results}
\begin{figure}[h!]
  \caption{PRISMA diagram for search publication retrieval}
  \includegraphics[width=12.4cm]{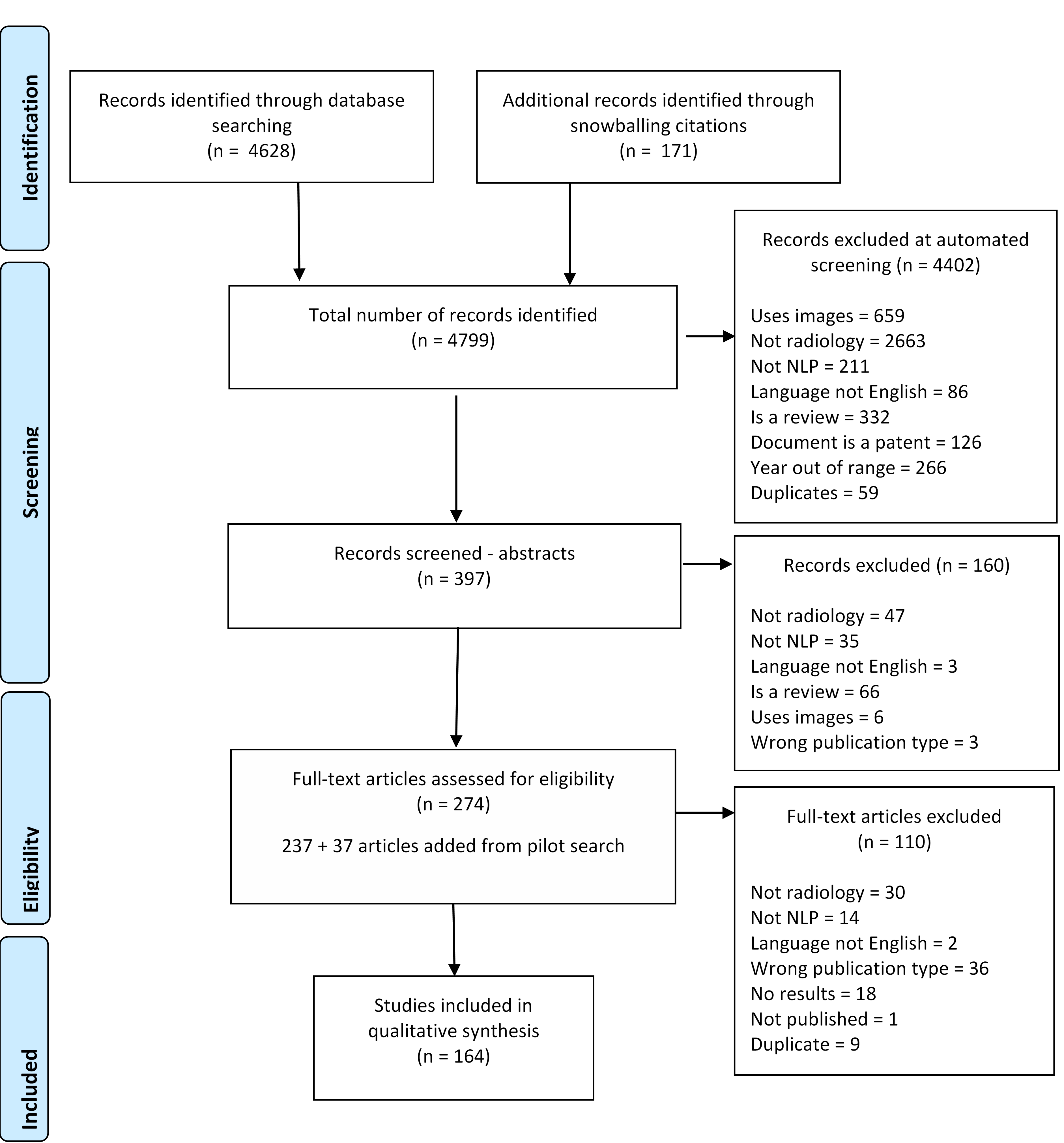}
   \label{fig:1}
\end{figure}

The literature search yielded 4,799 possibly relevant publications from which our automated exclusion process removed 4,402, and during both our screening processes, 233 were removed, leaving 164 publications. See Figure \ref{fig:1} for details of exclusions at each step. 
%Figure 1

\subsection*{General Characteristics}

 2015 and 2016 saw similar numbers of publications retrieved (22 and 21 respectively) with the volume increasing almost three-fold in 2019 (55), noting 2019 only covers 10 months (Figure \ref{fig:2}).  Imaging modality (Table \ref{tab:t3}) varied considerably and 38 studies used reports from multiple modalities. Of studies focusing on a single modality, the most featured were CT scans (36) followed by MRI (16), X-Ray (18), Mammogram (5) and Ultrasound (4). Forty-seven studies did not specifying scan modality. For the study samples (Table \ref{tab:t4}), 33 papers specified that they used consecutive patient images, 38 used non-consecutive image sampling and 93 did not clearly specify their sampling strategy.  The anatomical regions for scans varied (Table \ref{tab:t5}) with mixed being the highest followed by Thorax  and Head/neck.   Disease categories are presented in Table \ref{tab:t6} with the largest disease category being Oncology.  The majority of reports were in English (142) and a small number in other languages e.g., Chinese (5), Spanish (4), German (3) (Table \ref{tab:t7}).
%Table 3 & 4
\begin{table}[h!]
\begin{minipage}[t]{0.45\linewidth}\centering
\caption{Scan modality }
  \begin{tabular}{ll}
    \hline
    Scan Modality &	No. Studies\\
    \hline
Multiple Modalities &	38\\
MRI	& 16\\
CT	& 36\\
X-Ray &	18\\
Mammogram &	5\\
Ultrasound &	4\\
Not specified &	47\\
\hline
TOTAL	& 164\\
  \end{tabular}
  \label{tab:t3}
\end{minipage}
\hspace{0.5cm}
\begin{minipage}[t]{0.45\linewidth}

\caption{Image sampling method }
  \begin{tabular}{ll}
    \hline
   Sampling Method	No. Studies\\
   \hline
Consecutive Images	& 33\\
Non-Consecutive Images &	38\\
Not specified 	& 93\\
\hline
TOTAL &	164\\
  \end{tabular}
  \label{tab:t4}
\end{minipage}
\end{table}
%Table 5 & 6
\begin{table}[h!]
\begin{minipage}[t]{0.45\linewidth}\centering
\caption{Anatomical region scanned }
  \begin{tabular}{ll}
    \hline
   Anatomical Region &	No. Studies \\
   \hline
Mixed &	45 \\
Thorax &	31\\
Head/Neck &	25\\
Abdomen &	15\\
Breast &	15\\
Extremities	 & 8\\
Spine &	5\\
Other &	1\\
Unspecified	 & 19\\
\hline
TOTAL &	164
  \end{tabular}
  \label{tab:t5}
\end{minipage}
\hspace{0.5cm}
\begin{minipage}[t]{0.45\linewidth}

\caption{Disease category }
  \begin{tabular}{ll}
    \hline
    Disease Category &	No. Studies \\
    \hline
Not specific disease related	& 40\\
Oncology &	39\\
Various	& 20\\
Musculoskeletal	& 10\\
Cerebrovascular &	13\\
Other &	13\\
Respiratory &	10\\
Trauma &	7\\
Cardiovascular &	6\\
Gastrointestinal &	3\\
Hepatobiliary &	2\\
Genitourinary &	1\\
\hline
TOTAL &	164\\
  \end{tabular}
  \label{tab:t6}
\end{minipage}
\end{table}
%Table 7
\begin{table}[h!]
\caption{Radiology report language}
  \begin{tabular}{ll}
    \hline
   Report Language &	No. Studies\\
   \hline
English	& 142\\
Chinese	& 5\\
Spanish	& 4\\
German	& 3\\
Italian	& 2\\
French	& 2\\
Hebrew	& 1\\
Polish	& 1\\
Brazilian Portuguese	& 1\\
Unspecified	& 3\\
\hline
TOTAL &	164\\
  \end{tabular}
  \label{tab:t7}
\end{table}
%Table 8

\begin{table}[h!]
\caption{Clinical Application Category by Technical Objective}
  \begin{tabular}{lllll}
    \hline
    \multicolumn{1}{p{2cm}}{\raggedright Application Category} & \multicolumn{1}{p{2cm}}{\raggedright 	Information Extraction (n=81)} & \multicolumn{1}{p{1.5cm}}{\raggedright 	Report/ Sentence Classification (n=73) } & \multicolumn{1}{p{1.5cm}}{\raggedright 	Lexicon/ Ontology Discovery (n=9)} & \multicolumn{1}{p{1cm}}{\raggedright 	Clustering (n=1)}\\
 \hline
Disease Information \& Classification &	14	&31	-&	-\\
Diagnostic Surveillance&	28 & 	17&	-&	-\\
Quality Compliance&	7&	14&	-&	-\\
Cohort-Epid.&	6&	10&	-&	-\\
Language Discovery \& Knowledge&	13&	4&	9&	1\\
Technical NLP&	6&	4&	-&	-\\
  \end{tabular}
  \label{tab:t8}
\end{table}
\subsection*{Clinical Application Categories}

\begin{figure}[h!]
  \caption{Clinical application of publication by year}
  \includegraphics[width=\textwidth]{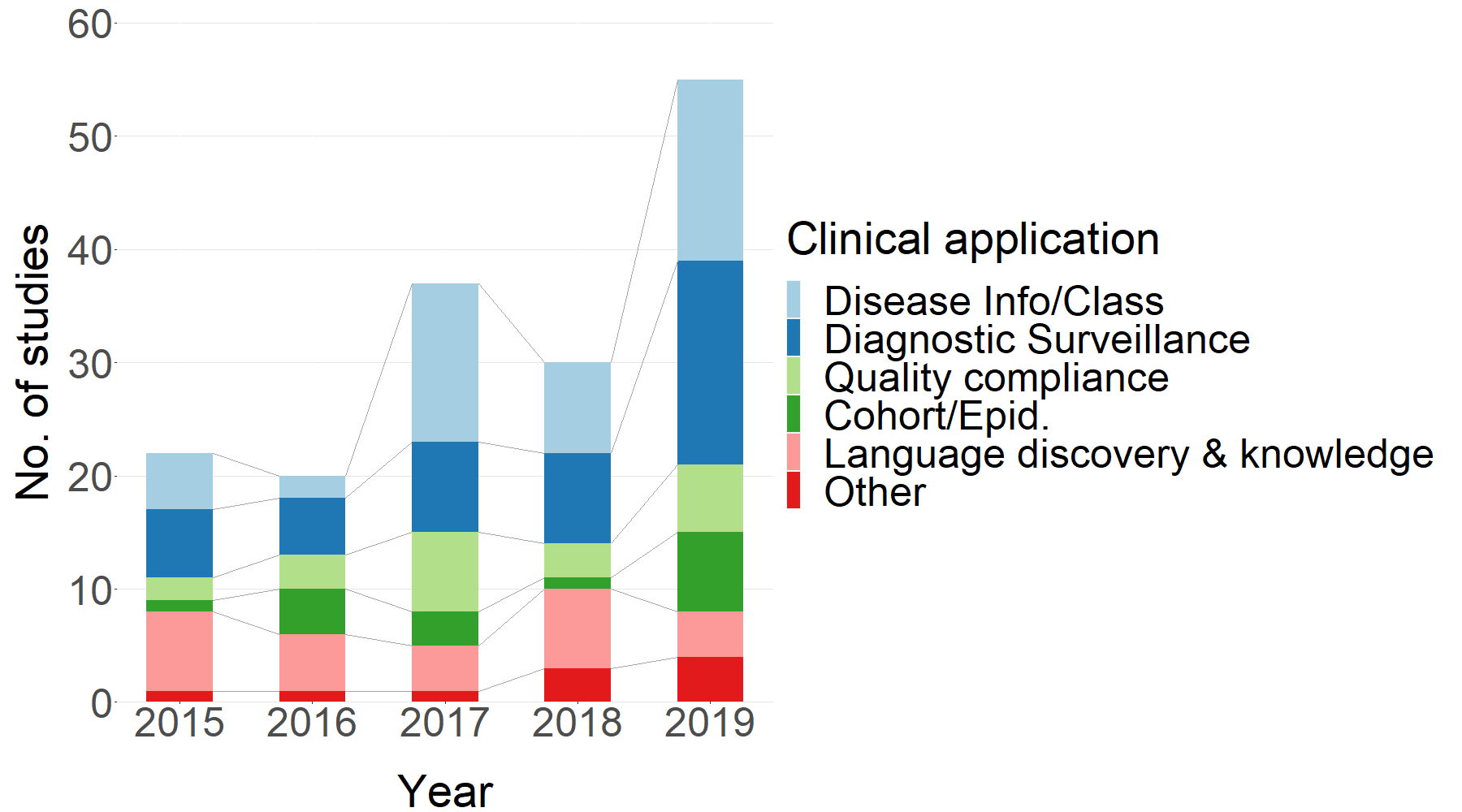}
   \label{fig:2}
\end{figure}

%Compared to 67 publications retrieved in the earlier review  of  \citep{pons_natural_2016}, we retrieved 164 publications. 
In synthesis of the literature each publication was classified by the primary clinical purpose. Pons'  work in 2016 categorised publications into 5 broad categories: Diagnostic Surveillance, Cohort Building for Epidemiological Studies, Query-based Case Retrieval, Quality Assessment of Radiological Practice and Clinical Support Services. We found some changes in this categorisation schema and our categorisation consisted of six categories: \textit{Diagnostic Surveillance, Disease information and classification, Quality Compliance, Cohort/Epidemiology, Language Discovery and Knowledge Structure, Technical NLP}.   The main difference is we found no evidence for a category of \textit{Clinical Support Services} which described applications that had been integrated into the workflow to assist. Despite the increase in the number of publications, very few were in clinical use with more focus on the category of \textit{Disease Information and Classification}. We describe each clinical application area in more detail below and where applicable how our categories differ from the earlier findings. A listing of all publications and their corresponding clinical application category can be %found in Additional File 1.EDITED for ARXIV 
made available on request.
Table \ref{tab:t8} shows the clinical application category by the technical classification and Figure \ref{fig:2} shows the breakdown of clinical application category by publication year. There were more publications in 2019 compared with 2015 for all categories except Language Discovery \& Knowledge Structure, which fell by $\approx$ 25\% (Figure 2).

% , particularly \textit{Diagnostic Surveillance} and \textit{Quality Compliance}.  However,  

% Their category of \textit{cohort for epidemiological studies} is the same as our combined category of \textit{Cohort/epidemiology}, but we treated these studies slightly differently in attempting to differentiate which papers described methods for creating cohorts for research purposes, and those which also reported the outcomes of an epidemiological analysis.  

% Each publication was classified by the primary clinical purpose into one of six categories: Diagnostic Surveillance, Disease information and classification, Quality Compliance, Cohort/Epidemiology, Language Discovery and Knowledge Structure, Other. We describe each clinical area in more detail below and a listing of all publications and their corresponding clinical application category can be found in Additional File 1.

%Figure 2

\subsubsection*{Diagnostic Surveillance}  
A large proportion of studies in this category focused on extracting disease information for patient or disease surveillance e.g. investigating tumour characteristics \citep{peng_self-attention_2019,bozkurt_automated_2019}; changes over time \citep{hassanpour_characterization_2017}  and worsening/progression or improvement/response to treatment \citep{kehl_assessment_2019,chen_integrating_2018}; identifying correct anatomical labels \citep{cotik_spanish_2018}; organ measurements and temporality \citep{sevenster_natural_2015-1}. Studies also investigated pairing measurements between reports \citep{sevenster_natural_2015} and linking reports to monitoring changes through providing an integrated view of consecutive examinations \citep{oberkampf_semantic_2016}.  Studies focused specifically on breast imaging findings investigating aspects, such as BI-RADS MRI descriptors (shape, size, margin) and final assessment categories (benign, malignant etc.) e.g., \citep{liu_automatic_2019,gupta_automatic_2018,castro_automated_2017,short_comprehensive_2019,lacson_assessing_2017,lacson_evaluation_2015}. Studies focused on tumour information e.g., for liver \citep{yim_tumor_2016-1} and hepatocellular carcinoma (HPC) \citep{yim_classifying_2017,yim_tumor_2016}  and one study on extracting information relevant for structuring subdural haematoma characteristics in reports \citep{pruitt_natural_2019}.

Studies in this category also investigated incidental findings including on lung imaging \citep{farjah_automated_2016,karunakaran_closing_2017,tan_comparison_2018}, with \citep{farjah_automated_2016}  additionally extracting the nodule size; for trauma patients \citep{trivedi_identifying_2019}; and looking for silent brain infarction and white matter disease \citep{fu_natural_2019}. Other studies focused on prioritising/triaging reports, detecting follow-up recommendations, and linking a follow-up exam to the initial recommendation report, or bio-surveillance of infectious conditions, such as invasive mould disease.

\subsubsection*{Disease Information and Classification}
\textit{Disease Information and Classification} publications use reports to identify information that may be aggregated according to classification systems.  These publications focused solely on classifying a disease occurrence or extracting information about a disease with no focus on the overall clinical application.  This category was not found in Pons' work. Methods considered  a range of conditions including intracranial haemorrhage \citep{jnawali_automatic_2019,banerjee_intelligent_2017}, aneurysms \citep{klos_automatic_2018}, brain metastases \citep{deshmukh_semi-supervised_2019}, ischaemic stroke \citep{kim_natural_2019,garg_automating_2019}, and several classified on types and severity of conditions e.g., \citep{deshmukh_semi-supervised_2019,shin_classification_2017,wheater_validated_2019,gorinski_named_2019,alex_text_2019}. Studies focused on breast imaging considered aspects such as predicting lesion malignancy from BI-RADS descriptors \citep{bozkurt_using_2016}, breast cancer subtypes \citep{patel_correlating_2017}, and extracting or inferring BI-RADS categories, such as \citep{banerjee_automatic_2019,miao_extraction_2018}. Two studies focused on abdominal images and hepatocellular carcinoma (HCC) staging and CLIP scoring.  Chest imaging reports were used to detect pulmonary embolism e.g., \citep{dunne_effect_2015,banerjee_comparative_2019,chen_deep_2017}, bacterial pneumonia \citep{meystre_enhancing_2017}, and Lungs-RADS categories \citep{beyer_automatic_2017}. Functional imaging was also included, such as echocardiograms, extracting measurements to evaluate heart failure, including left ventricular ejection fractions (LVEF). Other studies investigated classification of fractures and abnormalities and the prediction of ICD codes from imaging reports.

\subsubsection*{Language Discovery and Knowledge Structure}
\textit{Language Discovery and Knowledge Structure} publications investigate the structure of language in reports and how this might be optimised to facilitate decision support and communication. Pons et al. reported on applications of \textit{Query-based retrieval} which has similarities to \textit{Language Discovery and Knowledge Structure} but it is not the same.  Their category contains studies that retrieve cases and conditions that are not predefined and in some instances could be used for research purposes or are motivated for educational purposes.  Our category is broader and encompasses papers that investigated different aspects of language including variability, complexity simplification and normalising to support extraction and classification tasks.  
%In addition, we found a number of studies that used ontology based approaches to capture and visualise knowledge supporting clinicians not only to retrieve reports but in visualising the knowledge to support decision making.

 Studies focus on exploring lexicon coverage and methods to support language simplification for patients looking at sources, such as the consumer health vocabulary \citep{qenam_text_2017} and the French lexical network (JDM) \citep{lafourcade_radiological_2017}. Other works studied the variability and complexity of report language comparing free-text and structured reports and radiologists.  Also investigated was how ontologies and lexicons could be combined with other NLP methods to represent knowledge that can support clinicians. This work included improving report reading efficiency \citep{hong_investigation_2015}; finding similar reports \citep{comelli_ontology-based_2015}; normalising phrases to support classification and extraction tasks, such as entity recognition in Spanish reports \citep{cotik_approach_2015}; imputing semantic classes for labelling \citep{johnson_method_2015}, supporting search \citep{mujjiga_identifying_2019} or to discover semantic relations \citep{lafourcade_semantic_2016}.

\subsubsection*{Quality and Compliance}
\textit{Quality and Compliance} publications use reports to assess the quality and safety of practice and reports similar to Pons' category. Works considered how patient indications for scans adhered to guidance e.g., \citep{shelmerdine_automated_2019,mabotuwana_determining_2018,dalal_determining_2020,bobbin_focal_2017,kwan_follow_2019,mabotuwana_improving_2018} or protocol selection \citep{brown_natural_2017,trivedi_automatic_2018,zhang_development_2018,brown_using_2018,yan_yield_2016} or the impact of guideline changes on practice, such as \citep{kang_natural_2019}. Also investigated was diagnostic utilisation and yield, based on clinicians or on patients, which can be useful for hospital planning and for clinicians to study their work patterns e.g.\citep{brown_natural_2019}. Other studies in this category looked at specific aspects of quality, such as, classification for long bone fractures to support quality improvement in paediatric medicine \citep{grundmeier_identification_2016}, automatic identification of reports that have critical findings for auditing purposes \citep{heilbrun_feasibility_2019}, deriving a query-based quality measure to compare structured and free-text report variability \citep{maros_objective_2018}, and \citep{minn_improving_2015} who describe a method to fix errors in gender or laterality in a report.

\subsubsection*{Cohort and Epidemiology} 

 This category is similar to Pons' earlier review  but we treated the studies in this category slightly attempting to differentiate which papers described methods for creating cohorts for research purposes, and those which also reported the outcomes of an epidemiological analysis.
Ten studies use NLP to create specific cohorts for research purposes and six reported the performance of their tools. Out of these papers, the majority (n=8) created cohorts for specific medical conditions including fatty liver disease  \citep{goldshtein_identification_2020,redman_accurate_2017} hepatocellular cancer \citep{sada_validation_2016}, ureteric stones \citep{li_natural_2019}, vertebral facture \citep{tan_surrogate-guided_2019}, traumatic brain injury \citep{yadav_automated_2016,mahan_tbiextractor_2019}, and leptomeningeal disease secondary to metastatic breast cancer \citep{brizzi_natural_2019}. Five papers identified cohorts focused on particular radiology findings including ground glass opacities (GGO) \citep{van_haren_ground_2019}, cerebral microbleeds (CMB) \citep{noorbakhsh-sabet_racial_2018}, pulmonary nodules \citep{gould_recent_2015}, \citep{huhdanpaa_using_2018}, changes in the spine correlated to back pain \citep{bates_classification_2016} and identifying radiological evidence of people having suffered a fall. One paper focused on identifying abnormalities of specific anatomical regions of the ear within an audiology imaging database \citep{masino_temporal_2016} and another paper aimed to create a cohort of people with any rare disease (within existing ontologies -  Orphanet Rare Disease Ontology ORDO and Radiology Gamuts Ontology RGO). Lastly, one paper took a different approach of screening reports to create a cohort of people with contraindications for MRI, seeking to prevent iatrogenic events. 
 Amongst the epidemiology studies there were various analytical aims, but they primarily focused on estimating the prevalence or incidence of conditions or imaging findings and looking for associations of these conditions/findings with specific population demographics, associated factors or comorbidities. The focus of one study differed in that it applied NLP to healthcare evaluation, investigating the association of palliative care consultations and measures of high-quality end-of-life (EOL) care \citep{brizzi_natural_2019}.
\subsubsection*{Technical NLP}
This category is for publications that have a primary technical aim that is not focused on radiology report outcome, e.g. detecting negation in reports, spelling correction \citep{zech_detecting_2019}, fact checking \citep{zhang_optimizing_2019,steinkamp_toward_2019} methods for sample selection, crowd source annotation \citep{cocos_crowd_2017}.  This category did not occur in Pons' earlier review.

%%%%%%%ORIGINAL TEXT
% Diagnostic Surveillance publications use reports for surveillance of disease at a population or individual level. Disease Information and Classification publications use reports to identify information that may be aggregated according to classification systems.  Language Discovery and Knowledge Structure publications investigate the structure of language in reports and how this might be optimised to facilitate decision support and communication. Quality and Compliance publications use reports to assess the quality and safety of practice and reports. Cohort and Epidemiology publications use reports to create cohorts and carrying out research studies.  Other publication category use reports for purposes not fitting in the previous categories, usually with a primary technical aim, e.g. detecting negation in reports. 

\subsection*{NLP Methods in Use}

NLP methods capture the different techniques an author applied broken down into rules, machine learning methods, deep learning, ontologies, lexicons and word embeddings.  We discriminate machine learning from deep learning, using the former to represent traditional machine learning methods.

\begin{figure}[h!]
  \caption{NLP method breakdown}
  \includegraphics[width=8.5cm]{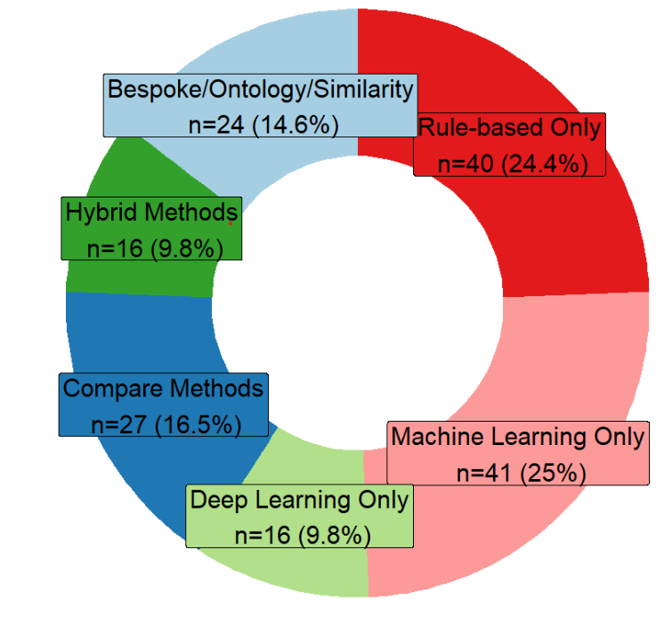}
   \label{fig:3}
\end{figure}

\begin{figure}[h!]
  \caption{NLP method by year}
  \includegraphics[width=\textwidth]{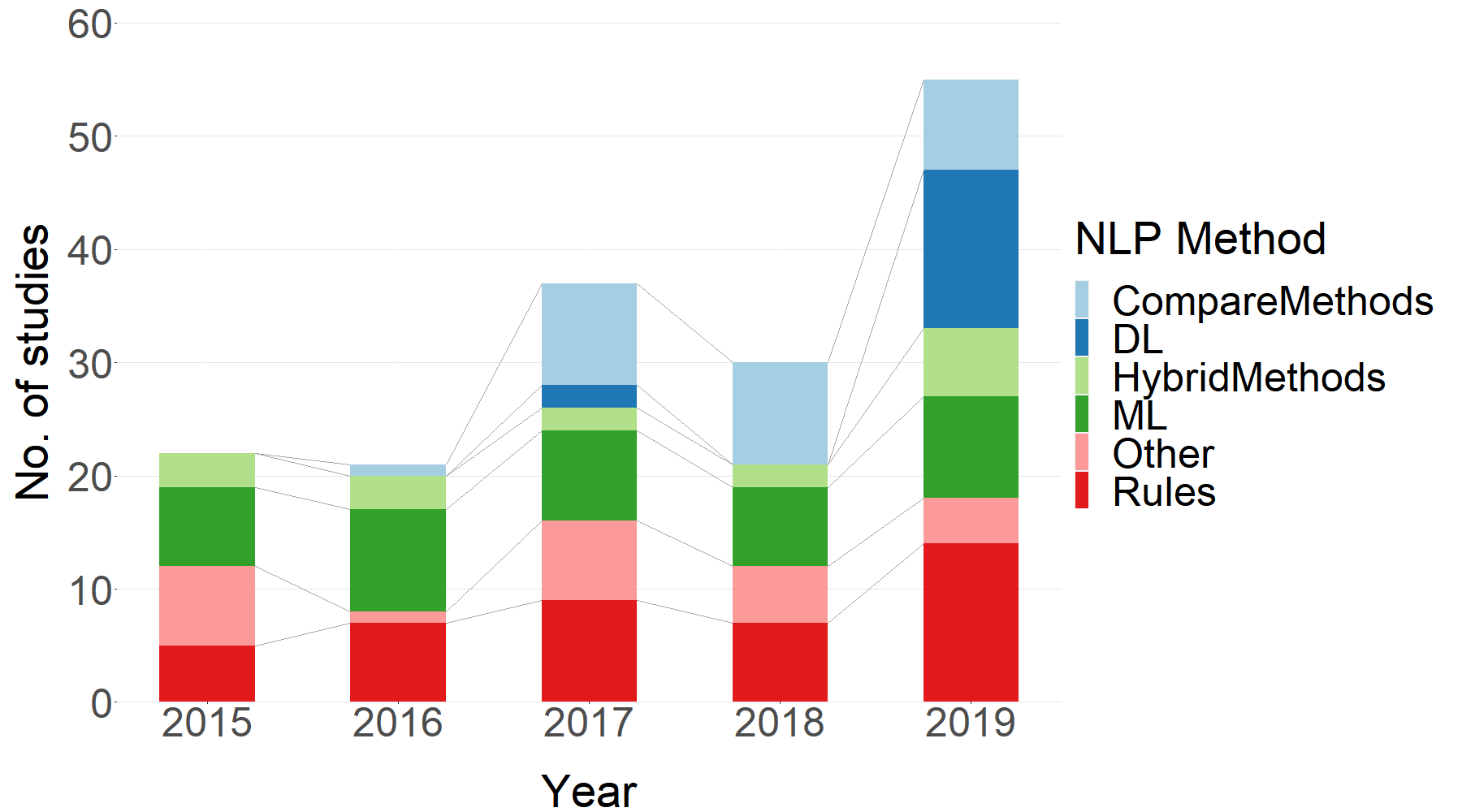}
   \label{fig:4}
\end{figure}

Over half of the studies only applied one type of NLP method and just over a quarter of the studies compared or combined methods in hybrid approaches. The remaining studies either used a bespoke proprietary system or focus on building ontologies or similarity measures (Figure \ref{fig:3}).  Rule-based method use remains almost constant across the period, whereas use of machine learning decreases and deep learning methods rises, from five publications in 2017 to twenty-four publications in 2019 (Figure \ref{fig:4}).  
%Figure 3
%Figure 4
%Figure 4

%Table 9
\begin{table}[h!]
    \caption{Breakdown of NLP Method}
    \begin{tabular}{l|l||l|l}
         ML (n=74) &	No studies &	Deep Learning (n=36) &	No studies\\
         \hline
SVM &	34	& RNN variants &	14\\
Logistic Regression	& 23 &	CNN &	10\\
Random Forest &	18 &	Other &	5\\
Naïve Bayes	& 17 &	Compare CNN, RNN &	4\\
Maximum Entropy &	7 &	Combine CNN+RNN &	3\\
Decision Trees &	4 & &	\\	
    \end{tabular}
      \label{tab:nlp_break}
\end{table}

A variety of machine classifier algorithms were used, with SVM and Logistic Regression being the most common (Table \ref{tab:nlp_break}).  Recurrent Neural Networks (RNN) variants were the most common type of deep learning architectures.  RNN methods were split between long short-term memory (LSTM) and bidirectional-LSTM (Bi-LSTM), bi-directional gated recurrent unit (Bi-GRU), and standard RNN approaches. Four of these studies additionally added a Conditional Random Field (CRF) for the final label generation step.   Convolutional Neural Networks (CNN) were the second most common architecture explored. Eight studies additionally used an attention mechanism as part of their deep learning architecture.  Other neural approaches included feed-forward neural networks, fully connected neural networks and a proprietary neural system IBM Watson \citep{trivedi_automatic_2018} and Snorkel \citep{ratner_snorkel_2018}.   Several studies proposed combined architectures, such as \citep{zhu_context-driven_2019,short_comprehensive_2019}.
\subsection*{NLP Method Features}
Most rule-based and machine classifying approaches used features based on bag-of-words, part-of-speech, term frequency, and phrases with only two studies alternatively using word embeddings. Three studies use feature engineering with deep learning rather than word embeddings. Thirty-three studies use domain-knowledge to support building features for their methods, such as developing lexicons or selecting terms and phrases.
Comparison of embedding methods is difficult as many studies did not describe their embedding method.  Of those that did, Word2Vec \citep{mikolov_efficient_2013} was the most popular (n=19), followed by GLOVE embeddings \citep{pennington_glove_2014} (n=6), FastText \citep{mikolov_advances_2018} (n=3), ELMo \citep{peters_deep_2018} (n=1) and BERT \citep{devlin_bert_2018} (n=1).   
Ontologies or lexicon look-ups are used in 100 studies; however, even though publications increase over the period in real terms, 20\% fewer studies employ the use of ontologies or lexicons in 2019 compared to 2015.  The most widely used resources were UMLS \citep{national_library_of_medicine_unified_2021} (n=15), Radlex \citep{rsna_radlex_2021} (n=20), SNOMED-CT \citep{national_library_of_medicine_snomed_2021} (n=14). Most studies used these as features for normalising words and phrases for classification, but this was mainly those using rule-based or machine learning classifiers with only six studies using ontologies as input to their deep learning architecture.  Three of those investigated how existing ontologies can be combined with word embeddings to create domain-specific mappings, with authors pointing to this avoiding the need for large amounts of annotated data.   Other approaches looked to extend existing medical resources using a frequent phrases approach, e.g. \citep{bulu_proposing_2018}. Works also used the derived concepts and relations visualising these to support activities, such as report reading and report querying (e.g. \citep{hassanpour_unsupervised_2016,zhao_using_2018}) 
\subsection*{Annotation and Inter-Annotator Agreement}
Eighty-nine studies used at least two annotators, 75 did not specify any annotation details, and only one study used a single annotator.  Whilst 69 studies use a domain expert for annotation (a clinician or radiologist) only 56 studies report the inter-annotator agreement. Some studies mention annotation but do not report on agreement or annotators.  Inter-annotator agreement values for Kappa range from 0.43 to perfect agreement at 1.  Whilst most studies reported agreement by Cohen's Kappa \citep{cohen_coefficient_1960} some reported precision, and percent agreement.
Studies reported annotation data sizes differently, e.g., on the sentence or patient level.  Studies also considered ground truth labels from coding schemes such as ICD or BI-RADS categories as annotated data.  Of studies which detailed human annotation at the radiology report level, only 45 specified inter-annotator agreement and/or the number of annotators. Annotated report numbers for these studies varies with 15 papers having annotated less than 500, 12 having annotated between 500 and less than 1,000, 15 between 1,000 and less than 3,000, and 3 between 4,000 and 8,288 reports.
\subsection*{Data Sources and Availability}
Only 14 studies reported that their data is available, and 15 studies reported that their code is available.  Most studies sourced their data from medical institutions, a number of studies did not specify where their data was from, and some studies used publicly available datasets: MIMIC-III (n=5), MIMIC-II (n=1), MIMIC-CXR (n=1); Radcore (n=5) or STRIDE (n=2).   Four studies used combined electronic health records such as clinical notes or pathology reports.
%Table 10
\begin{table}[h!]
     \caption{NLP Method by data size properties, minimum data size, maximum data size and median value, studies reporting in numbers of radiology reports}
    \begin{tabular}{l|l|l|l}
         NLP Method &	Min Size &	Max Size &	Median  \\
         \hline
         Compare Methods &	513	& 2,167,445	& 2,845 \\
         Hybrid Methods	& 40 &	34,926 &	918 \\
         Deep Learning (Only) &	120 &	1,567,581 &	5,000 \\
         Machine Learning (Only) &	101	& 2,977,739	& 2,531 \\
         Rules (Only)& 	31 &	10,000,000 &	8,000 \\
         Other	& 25 &	12,377,743 &	10,000 \\
    \end{tabular}
   
    \label{tab:nlpmethod_size}
\end{table}
%Table 11
\begin{table}[h!]
     \caption{Grouped data size and number of studies in each group, only for studies reporting in numbers of radiology reports}
    \begin{tabular}{l|l}
        Data Size  Group &	No. Studies (\%) \\
        \hline
\textless 200 &	9  (6.7) \\
200 \textless  500 &	6 (4.4)\\
500 \textless  1,000	& 18 (13.3)\\
1,000 \textless  2,000 &	17 (12.6)\\
2,000 \textless  5,000 &	17 (12.6)\\
5,000 \textless  10,000 &	12 (8.9)\\
10,000+	 & 53 (39.3)\\
Unspecified  &	3 (2.2)\\
    \end{tabular}
    \label{tab:group_data}
\end{table}

Reporting on data size and splits differed across studies with some not giving exact data sizes and others reporting numbers of sentences, patients, or mixed data sources rather than radiology reports. Data sizes for those reporting at the radiology report level is n=135 or 82.32\% of the studies (Table \ref{tab:nlpmethod_size}).  The biggest variation of data size by NLP Method is in studies that apply other methods or are rule-based.  Machine learning also varies in size; however, the median value is lower compared to rule-based methods. The median value for deep learning is considerably higher at 5,000 reports compared to machine learning or those that compare or create hybrid methods.
Of the studies reporting on radiology reports numbers, 39.3\% used over 10,000 reports and this increases to over 48\% using more than 5,000 reports. However, a small number of studies, 14\%, are using comparatively low numbers of radiology reports, less than 500 (Table \ref{tab:group_data}).
\subsection*{NLP Performance and Evaluation Measures}
\begin{figure}[h!]
  \caption{Application Category and NLP Method, Mean and Median Summaries.  Mean value is indicated by a vertical bar, the box shows error bars and the asterisk is the median value.}
  \includegraphics[width=\textwidth]{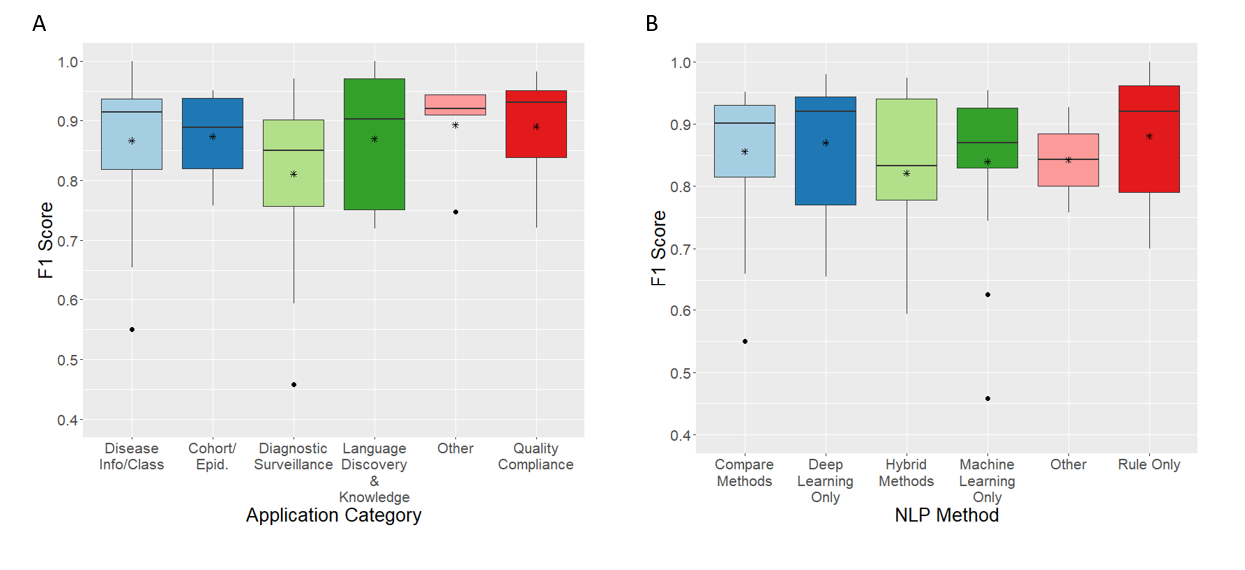}
   \label{fig:5}
\end{figure}

Performance metrics applied for evaluation of methods vary widely with authors using precision (positive predictive value (PPV)), recall (sensitivity), specificity, the area under the curve (AUC) or accuracy.
We observed a wide variety in evaluation methodology employed concerning test or validation datasets.  Different approaches were taken in generating splits for testing and validation, including k-fold cross-validation. Ninety-nine studies reported on training and test data splits, of which only 59 studies included a validation set.  Only 10 studies validated their algorithm using an external dataset from another institution, another modality, or a different patient population. 
The most widely used metrics for reporting performance were precision (PPV) and recall (sensitivity) reported in 47\% of studies. However, even though many studies compared methods and reported on the top-performing method, very few studies carried out significance testing on these comparisons.  Issues of heterogeneity make it difficult and unrealistic to compare performance between methods applied, hence, we use summary measures as a broad overview (Figure \ref{fig:5}).  Performance reported varies, but both the mean and median values for the F1 score appear higher for methods using rule-based only or deep learning only methods. Whilst differences are less discernible between F1 scores for application areas, \textit{Diagnostic Surveillance} looks on average lower than other categories. 
%Figure 5
\section*{Discussion and Future Directions}

Our work shows there has been a considerable increase in the number of publications using NLP on radiology reports over the recent time period.  Compared to 67 publications retrieved in the earlier review  of  \citep{pons_natural_2016}, we retrieved 164 publications.  In this section we discuss and offer some insight into the observations and trends of how NLP is being applied to radiology and make some recommendations that may benefit the field going forward.

\subsection*{Clinical Applications and NLP Methods in Radiology}
The clinical applications of the publications is similar to the earlier review of Pons et al. but whilst we observe an increase in research output we also highlight that there appears to be even less focus on clinical application compared to their review.  Like many other fields applying NLP the use of deep learning has  increased, with RNN architectures being the most popular. This is also observed in a review of NLP in clinical text\citep{wu_deep_2020}. However, although deep learning use increases, rules and traditional machine classifiers are still prevalent and often used as baselines to compare deep learning architectures against.  One reason for traditional methods remaining popular is their interpretability compared to deep learning models.  Understanding the features that drive a model prediction can support decision-making in the clinical domain but the complex layers of non-linear data transformations deep learning is composed of does not easily support transparency \citep{shickel_deep_2018}. This may also help explain why in synthesis of the literature we observed less focus on discussing clinical application and more emphasis on disease classification or information task only. Advances in interpretability of deep learning models are critical to its adoption in clinical practice.

Other challenges exist for deep learning such as only having access to small or imbalanced datasets.  Chen et al. \citep{chen_deep_2019} review deep learning methods within healthcare and point to these challenges resulting in poor performance but that these same datasets can perform well with traditional machine learning methods.  We found several studies highlight this and when data is scarce or datasets imbalanced, they introduced hybrid approaches of rules and deep learning to improve performance, particularly in the \textit{Diagnostic Surveillance} category. Yang et al. \citep{yang_towards_2018} observed rules performing better for some entity types, such as time and size, which are proportionally lower than some of the other entities in their train and test sets; hence they combine a bidirectional-LSTM and CRF with rules for entity recognition. Peng et al. \citep{peng_self-attention_2019} comment that combining rules and the neural architecture complement each other, with deep learning being more balanced between precision and recall, but the rule-based method having higher precision and lower recall. The authors reason that this provides better performance as rules can capture rare disease cases, particularly when multi-class labelling is needed, whilst deep learning architectures perform worse in instances with fewer data points.

In addition to its need for large-scale data, deep learning can be computationally costly.  The use of pre-trained models and embeddings may alleviate some of this burden.  Pre-trained models often only require fine-tuning, which can reduce computation cost.  Language comprehension pre-learned from other tasks can then be inherited from the parent models, meaning fewer domain-specific labelled examples may be needed \citep{wood_automated_2020}.  This use of pre-trained information also supports generalisability, e.g., \citep{banerjee_comparative_2019} show that their model trained on one dataset can generalise to other institutional datasets. 

Embedding use has increased which is expected with the application of deep learning approaches but many rule-based and machine classifiers continue to use traditional count-based features, e.g., bag-of-words and n-grams. Recent evidence \citep{ong_machine_2020} suggests that the trend to continue to use feature engineering with traditional machine learning methods does produce better performance in radiology reports than using domain-specific word embeddings.  

Banerjee et al. \citep{banerjee_intelligent_2017} found that there was not much difference between a uni-gram approach and a Word2vec embedding, hypothesising this was due to their narrow domain, intracranial haemorrhage. However,  the NLP research field has seen a move towards bi-directional encoder representations from transformers (BERT) based embedding models not reflected in our analysis, with only one study using BERT generated embeddings \citep{deshmukh_semi-supervised_2019}.   Embeddings from BERT are thought to be superior as they can deliver better contextual representations and result in improved task performance. Whilst more publications since our review period have used BERT based embeddings with radiology reports e.g. \citep{wood_automated_2020,smit_combining_2020} not all outperform traditional methods \citep{grivas_not_2020}.  Recent evidence shows that embeddings generated by BERT fail to show a generalisable understanding of negation \citep{ettinger_what_2020}, an essential factor in interpreting radiology reports effectively.   Specialised BERT models have been introduced such as ClinicalBERT \citep{alsentzer_publicly_2019} or BlueBERT \citep{smit_combining_2020}.  BlueBERT has been shown to outperform ClinicalBERT when considering chest radiology \citep{smit_chexbert_2020} but more exploration of the performance gains versus the benefits of generalisability are needed for radiology text.  

All NLP models have in common that they need large amounts of labelled data for model training \citep{yasaka_deep_2018}.   Several studies \citep{percha_expanding_2018,tahmasebi_automatic_2019,banerjee_radiology_2018} explored combining word embeddings and ontologies to create domain-specific mappings, and they suggest this can avoid a need for large amounts of annotated data. Additionally, \citep{percha_expanding_2018,tahmasebi_automatic_2019} highlight that such combinations could boost coverage and performance compared to more conventional techniques for concept normalisation.  

The number of publications using medical lexical knowledge resources is still relatively low, even though a recent trend in the general NLP field is to enhance deep learning with external knowledge \citep{young_recent_2018}.  This was also observed by \citep{wu_deep_2020}, where only 18\% of the deep learning studies in their review utilised knowledge resources.    Although pre-training supports learning previously known facts it could introduce unwanted bias, hindering performance.  The inclusion of domain expertise through resources such as medical lexical knowledge may help reduce this unwanted bias \citep{wu_deep_2020}.  Exploration of how this domain expertise can be incorporated with deep learning architectures in future could improve the performance when having access to less labelled data.  

\subsection*{Task Knowledge}
Knowledge about the disease area of interest and how aspects of this disease are linguistically expressed is useful and could promote  better performing solutions.  Whilst  \citep{donnelly_using_2019} find high variability between radiologists, with metric values (e.g. number of syntactic, clinical terms based on ontology mapping) being significantly greater on free-text than structured reports, \citep{xie_introducing_2019} who look specifically at anatomical areas find less evidence for variability.  Zech et al. \citep{zech_natural_2018} suggest that the highly specialised nature of each imaging modality creates different sub-languages and the ability to discover these labels (i.e. disease mentions) reflects the consistency with which labels are referred to. For example, edema is referred to very consistently whereas other labels are not, such as infarction/ischaemic.   Understanding the language and the context of entity mentions could help promote novel ideas on how to solve problems more effectively. For example, \citep{yim_classifying_2017} discuss how the accuracy of predicting malignancy is affected by cues being outside their window of consideration and \citep{yim_classification_2018} observe problems of co-reference resolution within a report due to long-range dependencies.  Both these studies use traditional NLP approaches, but we observed novel neural architectures being proposed to improve performance in similar tasks specifically capturing long-range context and dependency learning, e.g., \citep{zhu_context-driven_2019,short_comprehensive_2019}.  This understanding requires close cooperation of healthcare professionals and data scientists, which is different to some other fields where more disconnection is present \citep{chen_deep_2019}.

\subsection*{Study Heterogeneity, a Need for Reporting Standards}
Most studies reviewed could be described as a proof-of-concept and not trialled in a clinical setting.  Pons et al. \citep{pons_natural_2016} hypothesised that a lack of clinical application may stem from uncertainty around minimal performance requirements hampering implementations, evidence-based practice requiring justification and transparency of decisions, and the inability to be able to compare to human performance as the human agreement is often an unknown.  These hypotheses are still valid, and we see little evidence that these problems are solved.

Human annotation is generally considered the gold standard at measuring human performance, and whilst many studies reported that they used annotated data, overall, reporting was inconsistent.  Steps were undertaken to measure inter-annotator agreement (IAA), but in many studies, this was not directly comparable to the evaluation undertaken of the NLP methods. 
The size of the data being used to draw experimental conclusions from is important and accurate reporting of these measures is essential to ensure reproducibility and comparison in further studies. Reporting on the training, test and validation splits was varied with some studies not giving details and not using held-out validation sets.

Most studies use retrospective data from single institutions but this can lead to a model over-fitting and, thus, not generalising well when applied in a new setting.  Overcoming the problem of data availability is challenging due to privacy and ethics concerns, but essential to ensure that performance of models can be investigated across institutions, modalities, and methods.    Availability of data would allow for agreed benchmarks to be developed within the field that algorithm improvements can be measured upon.  External validation of applied methods was extremely low, although, this is likely due to the availability of external datasets. Making code available would enable researchers to report how external systems perform on their data.  However, only 15 studies reported that their code is available.  To be able to compare systems there is a need for common datasets to be available to benchmark and compare systems against. 

Whilst reported figures in precision and recall generally look high more evidence is needed for accurate comparison to human performance.  A wide variety of performance measures were used, with some studies only reporting one measure, e.g., accuracy or F1 scores, with these likely representing the best performance obtained.  Individual studies are often not directly comparable for such measures, but none-the-less clarity and consistency in reporting is desirable.   Many studies making model comparisons did not carry out any significance testing for these comparisons.

The make the following recommendations to help move the field forward, enable more inter-study comparisons, and increase study reproducibility:
\begin{enumerate}
    \item Clarity in reporting study properties is required: (a) Data characteristics including size and the type of dataset should be detailed, e.g., the number of reports, sentences, patients, and if patients how many reports per patient. The training, test and validation data split should be evident, as should the source of the data. 
    (b) Annotation characteristics including the methodology to develop the annotation should be reported, e.g., annotation set size, annotator details, how many, expertise. 
    (c) Performance metrics should include a range of metrics: precision, recall, F1, accuracy and not just one overall value.
\item	Significance testing should be carried out when a comparison between methods is made.
\item	Data and code availability are encouraged. While making data available will often be challenging due to privacy concerns, researchers should make code available to enable inter-study comparisons and external validation of methods.
\item Common datasets should be used to benchmark and compare systems.
\end{enumerate}

\subsection*{Limitations of Study}
Publication search is subject to bias in search methods and it is likely that our search strategy did inevitably miss some publications. Whilst trying to be precise and objective during our review process some of the data collected and categorising publications into categories was difficult to agree on and was subjective. For example, many of the publications could have belonged to more than one category.  One of the reasons for this was how diverse in structure the content was which was in some ways reflected by the different domains papers were published in.  It is also possible that certain keywords were missed in recording data elements due to the reviewers own biases and research experience.

\section*{Conclusions}
This paper presents an systematic review of publications using NLP on radiology reports during the period 2015 to October 2019.  We show there has been substantial growth in the field particularly in researchers using deep learning methods. Whilst deep learning use has increased, as seen in NLP research in general, it faces challenges of lower performance when data is scarce or when labelled data is unavailable, and is not widely used in clinical practice perhaps due to the difficulties in interpretability of such models.  Traditional machine learning and rule-based methods are, therefore, still widely in use.  Exploration of domain expertise such as medial lexical knowledge must be explored further to enhance performance when data is scarce.  The clinical domain faces challenges due to privacy and ethics in sharing data but overcoming this would enable development of benchmarks to measure algorithm performance and test model robustness across institutions.  Common agreed datasets to compare performance of tools against would help support the community in inter-study comparisons and validation of systems.  The work we present here has the potential to inform researchers about applications of NLP to radiology and to lead to more reliable and responsible research in the domain.

\section*{Acknowledgements}%% if any
Not applicable

\section*{Funding}%% if any
This research was supported by the Alan Turing Institute, MRC, HDR-UK and the Chief Scientist Office. B.A.,A.C,D.D.,A.G. and C.G. have been supported by the Alan Turing Institute via Turing Fellowships (B.A,C.G.) and Turing project funding (ESPRC grant EP/N510129/1).  A.G. was also funded by a MRC Mental Health Data Pathfinder Award (MRC-MCPC17209).  H.W. is MRC/Rutherford Fellow HRD UK (MR/S004149/1).  H.D. is supported by HDR UK National Phemomics Resource Project. V.S-P. is supported by the HDR UK National Text Analytics Implementation Project.  W.W. is supported by a Scottish Senior Clinical Fellowship (CAF/17/01).

\section*{Abbreviations}%% if any
NLP - natural language processing\\
e.g. - example\\
ICD - international classification of diseases\\
BI-RADS - Breast Imaging-Reporting and Data System\\
IAA - inter-annotator agreement\\
No.  - number\\
UMLS - unified medical language system\\
ELMo - embeddings from Language Models\\
BERT - bidirectional encoder representations form transformers\\
SVM - support vector machine\\
CNN - convolutional neural network\\
LSTM - long short-term memory\\
Bi-LSTM - bi-directional long short-term memory\\
Bi-GRU - bi-directional gated recurrent unit\\
CRF - conditional random field\\
GLOVE - Global Vectors for Word Representation\\

%\section*{Availability of data and materials}%% if any
%All data generated or analysed during this study are included in this published article [and its supplementary information files]
%
%\section*{Ethics approval and consent to participate}%% if any
%Not applicable
%
%\section*{Competing interests}
%The authors declare that they have no competing interests.
%
%\section*{Consent for publication}%% if any
%Not applicable
%
%\section*{Authors' contributions}
%B.A, W.W. and H.W. conceptualised this study. D.D. carried out the search including automated filtering and designing meta-enriching steps. BA, AG, CG and RT advised on the automatic data collection method devised by DD. M.T.C.P, A.G., H.D. and D.D carried out the first stage review and A.C., E.D., V.S-P, M.T.C.P, A.G., H.D., B.A. and D.D. carried out the second-stage review. A.C. synthesised the data and wrote the main manuscript with contributions from all authors.  All authors read and approved the final manuscript.

\bibliographystyle{plainnat}
\bibliography{SysReviewRadiologyNLP}

\begin{thebibliography}{132}
\providecommand{\natexlab}[1]{#1}
\providecommand{\url}[1]{\texttt{#1}}
\expandafter\ifx\csname urlstyle\endcsname\relax
  \providecommand{\doi}[1]{doi: #1}\else
  \providecommand{\doi}{doi: \begingroup \urlstyle{rm}\Url}\fi

\bibitem[Alex et~al.(2019)Alex, Grover, Tobin, Sudlow, Mair, and
  Whiteley]{alex_text_2019}
Beatrice Alex, Claire Grover, Richard Tobin, Cathie Sudlow, Grant Mair, and
  William Whiteley.
\newblock Text mining brain imaging reports.
\newblock \emph{Journal of Biomedical Semantics}, 10\penalty0 (1):\penalty0 23,
  November 2019.
\newblock ISSN 2041-1480.
\newblock \doi{10.1186/s13326-019-0211-7}.
\newblock URL \url{https://doi.org/10.1186/s13326-019-0211-7}.

\bibitem[Alsentzer et~al.(2019)Alsentzer, Murphy, Boag, Weng, Jindi, Naumann,
  and McDermott]{alsentzer_publicly_2019}
Emily Alsentzer, John Murphy, William Boag, Wei-Hung Weng, Di~Jindi, Tristan
  Naumann, and Matthew McDermott.
\newblock Publicly {Available} {Clinical} {BERT} {Embeddings}.
\newblock In \emph{Proceedings of the 2nd {Clinical} {Natural} {Language}
  {Processing} {Workshop}}, pages 72--78, Minneapolis, Minnesota, USA, June
  2019. Association for Computational Linguistics.
\newblock \doi{10.18653/v1/W19-1909}.
\newblock URL \url{https://www.aclweb.org/anthology/W19-1909}.

\bibitem[Banerjee et~al.(2017)Banerjee, Madhavan, Goldman, and
  Rubin]{banerjee_intelligent_2017}
Imon Banerjee, Sriraman Madhavan, Roger~Eric Goldman, and Daniel~L. Rubin.
\newblock Intelligent {Word} {Embeddings} of {Free}-{Text} {Radiology}
  {Reports}.
\newblock \emph{AMIA Annual Symposium Proceedings}, pages 411--420, 2017.
\newblock ISSN 1942-597X.
\newblock URL \url{https://www.ncbi.nlm.nih.gov/pmc/articles/PMC5977573/}.

\bibitem[Banerjee et~al.(2018)Banerjee, Chen, Lungren, and
  Rubin]{banerjee_radiology_2018}
Imon Banerjee, Matthew~C. Chen, Matthew~P. Lungren, and Daniel~L. Rubin.
\newblock Radiology report annotation using intelligent word embeddings:
  {Applied} to multi-institutional chest {CT} cohort.
\newblock \emph{Journal of Biomedical Informatics}, 77:\penalty0 11--20,
  January 2018.
\newblock ISSN 1532-0464.
\newblock \doi{10.1016/j.jbi.2017.11.012}.
\newblock URL
  \url{http://www.sciencedirect.com/science/article/pii/S1532046417302575}.

\bibitem[Banerjee et~al.(2019{\natexlab{a}})Banerjee, Bozkurt, Alkim, Sagreiya,
  Kurian, and Rubin]{banerjee_automatic_2019}
Imon Banerjee, Selen Bozkurt, Emel Alkim, Hersh Sagreiya, Allison~W. Kurian,
  and Daniel~L. Rubin.
\newblock Automatic inference of {BI}-{RADS} final assessment categories from
  narrative mammography report findings.
\newblock \emph{Journal of Biomedical Informatics}, 92:\penalty0 103137, April
  2019{\natexlab{a}}.
\newblock ISSN 1532-0464.
\newblock \doi{10.1016/j.jbi.2019.103137}.
\newblock URL
  \url{http://www.sciencedirect.com/science/article/pii/S1532046419300553}.

\bibitem[Banerjee et~al.(2019{\natexlab{b}})Banerjee, Ling, Chen, Hasan,
  Langlotz, Moradzadeh, Chapman, Amrhein, Mong, Rubin, Farri, and
  Lungren]{banerjee_comparative_2019}
Imon Banerjee, Yuan Ling, Matthew~C. Chen, Sadid~A. Hasan, Curtis~P. Langlotz,
  Nathaniel Moradzadeh, Brian Chapman, Timothy Amrhein, David Mong, Daniel~L.
  Rubin, Oladimeji Farri, and Matthew~P. Lungren.
\newblock Comparative effectiveness of convolutional neural network ({CNN}) and
  recurrent neural network ({RNN}) architectures for radiology text report
  classification.
\newblock \emph{Artificial Intelligence in Medicine}, 97:\penalty0 79--88, June
  2019{\natexlab{b}}.
\newblock ISSN 0933-3657.
\newblock \doi{10.1016/j.artmed.2018.11.004}.
\newblock URL
  \url{http://www.sciencedirect.com/science/article/pii/S0933365717306255}.

\bibitem[Bates et~al.(2016)Bates, Fodeh, Brandt, and
  Womack]{bates_classification_2016}
Jonathan Bates, Samah~J. Fodeh, Cynthia~A. Brandt, and Julie~A. Womack.
\newblock Classification of radiology reports for falls in an {HIV} study
  cohort.
\newblock \emph{Journal of the American Medical Informatics Association},
  23\penalty0 (e1):\penalty0 e113--e117, April 2016.
\newblock ISSN 1067-5027.
\newblock \doi{10.1093/jamia/ocv155}.
\newblock URL \url{https://academic.oup.com/jamia/article/23/e1/e113/2379897}.

\bibitem[Beyer et~al.(2017)Beyer, McKee, Regis, McKee, Flacke, El~Saadawi, and
  Wald]{beyer_automatic_2017}
Sebastian~E. Beyer, Brady~J. McKee, Shawn~M. Regis, Andrea~B. McKee, Sebastian
  Flacke, Gilan El~Saadawi, and Christoph Wald.
\newblock Automatic {Lung}-{RADS}™ classification with a natural language
  processing system.
\newblock \emph{Journal of Thoracic Disease}, 9\penalty0 (9):\penalty0
  3114--3122, September 2017.
\newblock ISSN 2072-1439.
\newblock \doi{10.21037/jtd.2017.08.13}.
\newblock URL \url{https://www.ncbi.nlm.nih.gov/pmc/articles/PMC5708435/}.

\bibitem[Bobbin et~al.(2017)Bobbin, Ip, Sahni, Shinagare, and
  Khorasani]{bobbin_focal_2017}
Mark~D. Bobbin, Ivan~K. Ip, V.~Anik Sahni, Atul~B. Shinagare, and Ramin
  Khorasani.
\newblock Focal {Cystic} {Pancreatic} {Lesion} {Follow}-up {Recommendations}
  {After} {Publication} of {ACR} {White} {Paper} on {Managing} {Incidental}
  {Findings}.
\newblock \emph{Journal of the American College of Radiology}, 14\penalty0
  (6):\penalty0 757--764, June 2017.
\newblock ISSN 1546-1440.
\newblock \doi{10.1016/j.jacr.2017.01.044}.
\newblock URL
  \url{http://www.sciencedirect.com/science/article/pii/S1546144017301771}.

\bibitem[Bozkurt et~al.(2016)Bozkurt, Gimenez, Burnside, Gulkesen, and
  Rubin]{bozkurt_using_2016}
Selen Bozkurt, Francisco Gimenez, Elizabeth~S. Burnside, Kemal~H. Gulkesen, and
  Daniel~L. Rubin.
\newblock Using automatically extracted information from mammography reports
  for decision-support.
\newblock \emph{Journal of Biomedical Informatics}, 62:\penalty0 224--231,
  August 2016.
\newblock ISSN 1532-0464.
\newblock \doi{10.1016/j.jbi.2016.07.001}.
\newblock URL
  \url{http://www.sciencedirect.com/science/article/pii/S1532046416300557}.

\bibitem[Bozkurt et~al.(2019)Bozkurt, Alkim, Banerjee, and
  Rubin]{bozkurt_automated_2019}
Selen Bozkurt, Emel Alkim, Imon Banerjee, and Daniel~L. Rubin.
\newblock Automated {Detection} of {Measurements} and {Their} {Descriptors} in
  {Radiology} {Reports} {Using} a {Hybrid} {Natural} {Language} {Processing}
  {Algorithm}.
\newblock \emph{Journal of Digital Imaging}, 32\penalty0 (4):\penalty0
  544--553, August 2019.
\newblock ISSN 1618-727X.
\newblock \doi{10.1007/s10278-019-00237-9}.
\newblock URL \url{https://doi.org/10.1007/s10278-019-00237-9}.

\bibitem[Briscoe et~al.(2020)Briscoe, Bethel, and Rogers]{briscoe_conduct_2020}
Simon Briscoe, Alison Bethel, and Morwenna Rogers.
\newblock Conduct and reporting of citation searching in {Cochrane} systematic
  reviews: {A} cross-sectional study.
\newblock \emph{Research Synthesis Methods}, 11\penalty0 (2):\penalty0
  169--180, 2020.
\newblock ISSN 1759-2887.
\newblock \doi{10.1002/jrsm.1355}.
\newblock URL \url{https://onlinelibrary.wiley.com/doi/abs/10.1002/jrsm.1355}.

\bibitem[Brizzi et~al.(2019)Brizzi, Zupanc, Udelsman, Tulsky, Wright, Poort,
  and Lindvall]{brizzi_natural_2019}
Kate Brizzi, Sophia~N. Zupanc, Brooks~V. Udelsman, James~A. Tulsky, Alexi~A.
  Wright, Hanneke Poort, and Charlotta Lindvall.
\newblock Natural {Language} {Processing} to {Assess} {Palliative} {Care} and
  {End}-of-{Life} {Process} {Measures} in {Patients} {With} {Breast} {Cancer}
  {With} {Leptomeningeal} {Disease}.
\newblock \emph{American Journal of Hospice and Palliative Medicine},
  37\penalty0 (5):\penalty0 371--376, 2019.
\newblock \doi{https://doi.org/10.1177/1049909119885585}.
\newblock URL
  \url{https://journals.sagepub.com/doi/abs/10.1177/1049909119885585}.

\bibitem[Brown and Kachura(2019)]{brown_natural_2019}
A.~D. Brown and J.~R. Kachura.
\newblock Natural {Language} {Processing} of {Radiology} {Reports} in
  {Patients} {With} {Hepatocellular} {Carcinoma} to {Predict} {Radiology}
  {Resource} {Utilization}.
\newblock \emph{Journal of the American College of Radiology}, 16\penalty0
  (6):\penalty0 840--844, June 2019.
\newblock ISSN 1546-1440.
\newblock \doi{10.1016/j.jacr.2018.12.004}.
\newblock URL
  \url{http://www.sciencedirect.com/science/article/pii/S1546144018315539}.

\bibitem[Brown and Marotta(2017)]{brown_natural_2017}
Andrew~D. Brown and Thomas~R. Marotta.
\newblock A {Natural} {Language} {Processing}-based {Model} to {Automate} {MRI}
  {Brain} {Protocol} {Selection} and {Prioritization}.
\newblock \emph{Academic Radiology}, 24\penalty0 (2):\penalty0 160--166,
  February 2017.
\newblock ISSN 1076-6332.
\newblock \doi{10.1016/j.acra.2016.09.013}.
\newblock URL
  \url{http://www.sciencedirect.com/science/article/pii/S1076633216303270}.

\bibitem[Brown and Marotta(2018)]{brown_using_2018}
Andrew~D. Brown and Thomas~R. Marotta.
\newblock Using machine learning for sequence-level automated {MRI} protocol
  selection in neuroradiology.
\newblock \emph{Journal of the American Medical Informatics Association},
  25\penalty0 (5):\penalty0 568--571, May 2018.
\newblock ISSN 1067-5027.
\newblock \doi{10.1093/jamia/ocx125}.
\newblock URL \url{https://academic.oup.com/jamia/article/25/5/568/4569611}.

\bibitem[Bulu et~al.(2018)Bulu, Sippo, Lee, Burnside, and
  Rubin]{bulu_proposing_2018}
Hakan Bulu, Dorothy~A. Sippo, Janie~M. Lee, Elizabeth~S. Burnside, and
  Daniel~L. Rubin.
\newblock Proposing {New} {RadLex} {Terms} by {Analyzing} {Free}-{Text}
  {Mammography} {Reports}.
\newblock \emph{Journal of Digital Imaging}, 31\penalty0 (5):\penalty0
  596--603, October 2018.
\newblock ISSN 1618-727X.
\newblock \doi{10.1007/s10278-018-0064-0}.
\newblock URL \url{https://doi.org/10.1007/s10278-018-0064-0}.

\bibitem[Cai et~al.(2016)Cai, Giannopoulos, Yu, Kelil, Ripley, Kumamaru,
  Rybicki, and Mitsouras]{cai_natural_2016}
Tianrun Cai, Andreas~A. Giannopoulos, Sheng Yu, Tatiana Kelil, Beth Ripley,
  Kanako~K. Kumamaru, Frank~J. Rybicki, and Dimitrios Mitsouras.
\newblock Natural {Language} {Processing} {Technologies} in {Radiology}
  {Research} and {Clinical} {Applications}.
\newblock \emph{RadioGraphics}, 36\penalty0 (1):\penalty0 176--191, January
  2016.
\newblock ISSN 0271-5333.
\newblock \doi{10.1148/rg.2016150080}.
\newblock URL \url{https://pubs.rsna.org/doi/full/10.1148/rg.2016150080}.

\bibitem[Castro et~al.(2017)Castro, Tseytlin, Medvedeva, Mitchell, Visweswaran,
  Bekhuis, and Jacobson]{castro_automated_2017}
Sergio~M. Castro, Eugene Tseytlin, Olga Medvedeva, Kevin Mitchell, Shyam
  Visweswaran, Tanja Bekhuis, and Rebecca~S. Jacobson.
\newblock Automated annotation and classification of {BI}-{RADS} assessment
  from radiology reports.
\newblock \emph{Journal of Biomedical Informatics}, 69:\penalty0 177--187, May
  2017.
\newblock ISSN 1532-0464.
\newblock \doi{10.1016/j.jbi.2017.04.011}.
\newblock URL
  \url{http://www.sciencedirect.com/science/article/pii/S1532046417300813}.

\bibitem[Chen et~al.(2019)Chen, Liu, Kingsbury, Sohn, Storlie, Habermann,
  Naessens, Larson, and Liu]{chen_deep_2019}
David Chen, Sijia Liu, Paul Kingsbury, Sunghwan Sohn, Curtis~B. Storlie,
  Elizabeth~B. Habermann, James~M. Naessens, David~W. Larson, and Hongfang Liu.
\newblock Deep learning and alternative learning strategies for retrospective
  real-world clinical data.
\newblock \emph{npj Digital Medicine}, 2\penalty0 (1):\penalty0 1--5, May 2019.
\newblock ISSN 2398-6352.
\newblock \doi{10.1038/s41746-019-0122-0}.
\newblock URL \url{https://www.nature.com/articles/s41746-019-0122-0}.

\bibitem[Chen et~al.(2017)Chen, Ball, Yang, Moradzadeh, Chapman, Larson,
  Langlotz, Amrhein, and Lungren]{chen_deep_2017}
Matthew~C. Chen, Robyn~L. Ball, Lingyao Yang, Nathaniel Moradzadeh, Brian~E.
  Chapman, David~B. Larson, Curtis~P. Langlotz, Timothy~J. Amrhein, and
  Matthew~P. Lungren.
\newblock Deep {Learning} to {Classify} {Radiology} {Free}-{Text} {Reports}.
\newblock \emph{Radiology}, 286\penalty0 (3):\penalty0 845--852, November 2017.
\newblock ISSN 0033-8419.
\newblock \doi{10.1148/radiol.2017171115}.
\newblock URL \url{https://pubs.rsna.org/doi/full/10.1148/radiol.2017171115}.

\bibitem[Chen et~al.(2018)Chen, Zafar, Galperin-Aizenberg, and
  Cook]{chen_integrating_2018}
Po-Hao Chen, Hanna Zafar, Maya Galperin-Aizenberg, and Tessa Cook.
\newblock Integrating {Natural} {Language} {Processing} and {Machine}
  {Learning} {Algorithms} to {Categorize} {Oncologic} {Response} in {Radiology}
  {Reports}.
\newblock \emph{Journal of Digital Imaging}, 31\penalty0 (2):\penalty0
  178--184, April 2018.
\newblock ISSN 1618-727X.
\newblock \doi{10.1007/s10278-017-0027-x}.
\newblock URL \url{https://doi.org/10.1007/s10278-017-0027-x}.

\bibitem[Cocos et~al.(2017)Cocos, Qian, Callison-Burch, and
  Masino]{cocos_crowd_2017}
Anne Cocos, Ting Qian, Chris Callison-Burch, and Aaron~J. Masino.
\newblock Crowd control: {Effectively} utilizing unscreened crowd workers for
  biomedical data annotation.
\newblock \emph{Journal of Biomedical Informatics}, 69:\penalty0 86--92, May
  2017.
\newblock ISSN 1532-0464.
\newblock \doi{10.1016/j.jbi.2017.04.003}.
\newblock URL
  \url{http://www.sciencedirect.com/science/article/pii/S1532046417300746}.

\bibitem[Cohen(1960)]{cohen_coefficient_1960}
Jacob Cohen.
\newblock A {Coefficient} of {Agreement} for {Nominal} {Scales}.
\newblock \emph{Educational and Psychological Measurement}, 20\penalty0
  (1):\penalty0 37--46, April 1960.
\newblock ISSN 0013-1644.
\newblock \doi{10.1177/001316446002000104}.
\newblock URL \url{https://doi.org/10.1177/001316446002000104}.

\bibitem[Comelli et~al.(2015)Comelli, Agnello, and
  Vitabile]{comelli_ontology-based_2015}
A.~Comelli, L.~Agnello, and S.~Vitabile.
\newblock An ontology-based retrieval system for mammographic reports.
\newblock In \emph{2015 {IEEE} {Symposium} on {Computers} and {Communication}
  ({ISCC})}, pages 1001--1006, Larnaca, July 2015. IEEE.
\newblock \doi{10.1109/ISCC.2015.7405644}.

\bibitem[Cotik et~al.(2015)Cotik, Filippo, and Castano]{cotik_approach_2015}
Viviana Cotik, Dario Filippo, and Jose Castano.
\newblock An {Approach} for {Automatic} {Classification} of {Radiology}
  {Reports} in {Spanish}.
\newblock \emph{Studies in Health Technology and Informatics}, 216:\penalty0
  634--638, jan 2015.
\newblock ISSN 0926-9630, 1879-8365.
\newblock URL \url{https://europepmc.org/article/med/26262128}.

\bibitem[Cotik et~al.(2018)Cotik, Rodríguez, and Vivaldi]{cotik_spanish_2018}
Viviana Cotik, Horacio Rodríguez, and Jorge Vivaldi.
\newblock Spanish {Named} {Entity} {Recognition} in the {Biomedical} {Domain}.
\newblock In Juan~Antonio Lossio-Ventura, Denisse Muñante, and Hugo
  Alatrista-Salas, editors, \emph{Information {Management} and {Big} {Data}},
  volume 898 of \emph{Communications in {Computer} and {Information}
  {Science}}, pages 233--248, Lima, Peru, 2018. Springer International
  Publishing.
\newblock ISBN 978-3-030-11680-4.
\newblock \doi{10.1007/978-3-030-11680-4-23}.

\bibitem[Dalal et~al.(2020)Dalal, Hombal, Weng, Mankovich, Mabotuwana, Hall,
  Fuller, Lehnert, and Gunn]{dalal_determining_2020}
Sandeep Dalal, Vadiraj Hombal, Wei-Hung Weng, Gabe Mankovich, Thusitha
  Mabotuwana, Christopher~S. Hall, Joseph Fuller, Bruce~E. Lehnert, and
  Martin~L. Gunn.
\newblock Determining {Follow}-{Up} {Imaging} {Study} {Using} {Radiology}
  {Reports}.
\newblock \emph{Journal of Digital Imaging}, 33\penalty0 (1):\penalty0
  121--130, February 2020.
\newblock ISSN 1618-727X.
\newblock \doi{10.1007/s10278-019-00260-w}.
\newblock URL \url{https://doi.org/10.1007/s10278-019-00260-w}.

\bibitem[Deshmukh et~al.(2019)Deshmukh, Gumustop, Gauriau, Buch, Wright,
  Bridge, Naidu, Andriole, and Bizzo]{deshmukh_semi-supervised_2019}
Neil Deshmukh, Selin Gumustop, Romane Gauriau, Varun Buch, Bradley Wright,
  Christopher Bridge, Ram Naidu, Katherine Andriole, and Bernardo Bizzo.
\newblock Semi-{Supervised} {Natural} {Language} {Approach} for
  {Fine}-{Grained} {Classification} of {Medical} {Reports}.
\newblock \emph{arXiv:1910.13573 [cs.LG]}, November 2019.
\newblock URL \url{http://arxiv.org/abs/1910.13573}.

\bibitem[Devlin et~al.(2018)Devlin, Chang, Lee, and
  Toutanova]{devlin_bert_2018}
Jacob Devlin, Ming-Wei Chang, Kenton Lee, and Kristina Toutanova.
\newblock Bert: {Pre}-training of deep bidirectional transformers for language
  understanding.
\newblock \emph{arXiv preprint arXiv:1810.04805}, 2018.

\bibitem[Donnelly et~al.(2019)Donnelly, Grzeszczuk, Guimaraes, Zhang, and
  Bisset~III]{donnelly_using_2019}
Lane~F. Donnelly, Robert Grzeszczuk, Carolina~V. Guimaraes, Wei Zhang, and
  George~S. Bisset~III.
\newblock Using a {Natural} {Language} {Processing} and {Machine} {Learning}
  {Algorithm} {Program} to {Analyze} {Inter}-{Radiologist} {Report} {Style}
  {Variation} and {Compare} {Variation} {Between} {Radiologists} {When} {Using}
  {Highly} {Structured} {Versus} {More} {Free} {Text} {Reporting}.
\newblock \emph{Current Problems in Diagnostic Radiology}, 48\penalty0
  (6):\penalty0 524--530, November 2019.
\newblock ISSN 0363-0188.
\newblock \doi{10.1067/j.cpradiol.2018.09.005}.
\newblock URL
  \url{http://www.sciencedirect.com/science/article/pii/S0363018818302081}.

\bibitem[Dunne et~al.(2015)Dunne, Ip, Abbett, Gershanik, Raja, Hunsaker, and
  Khorasani]{dunne_effect_2015}
Ruth~M. Dunne, Ivan~K. Ip, Sarah Abbett, Esteban~F. Gershanik, Ali~S. Raja,
  Andetta Hunsaker, and Ramin Khorasani.
\newblock Effect of {Evidence}-based {Clinical} {Decision} {Support} on the
  {Use} and {Yield} of {CT} {Pulmonary} {Angiographic} {Imaging} in
  {Hospitalized} {Patients}.
\newblock \emph{Radiology}, 276\penalty0 (1):\penalty0 167--174, February 2015.
\newblock ISSN 0033-8419.
\newblock \doi{10.1148/radiol.15141208}.
\newblock URL \url{https://pubs.rsna.org/doi/full/10.1148/radiol.15141208}.

\bibitem[Ettinger(2020)]{ettinger_what_2020}
Allyson Ettinger.
\newblock What {BERT} {Is} {Not}: {Lessons} from a {New} {Suite} of
  {Psycholinguistic} {Diagnostics} for {Language} {Models}.
\newblock \emph{Transactions of the Association for Computational Linguistics},
  8:\penalty0 34--48, January 2020.
\newblock \doi{10.1162/tacl\_a\_00298}.
\newblock URL \url{https://doi.org/10.1162/tacl_a_00298}.

\bibitem[Farjah et~al.(2016)Farjah, Halgrim, Buist, Gould, Zeliadt, Loggers,
  and Carrell]{farjah_automated_2016}
Farhood Farjah, Scott Halgrim, Diana~S.M. Buist, Michael~K. Gould, Steven~B.
  Zeliadt, Elizabeth~T. Loggers, and David~S. Carrell.
\newblock An {Automated} {Method} for {Identifying} {Individuals} with a {Lung}
  {Nodule} {Can} {Be} {Feasibly} {Implemented} {Across} {Health} {Systems}.
\newblock \emph{eGEMs}, 4\penalty0 (1):\penalty0 1254, August 2016.
\newblock ISSN 2327-9214.
\newblock \doi{10.13063/2327-9214.1254}.
\newblock URL \url{https://www.ncbi.nlm.nih.gov/pmc/articles/PMC5013935/}.

\bibitem[Fleiss(1971)]{fleiss_measuring_1971}
Joseph~L. Fleiss.
\newblock Measuring nominal scale agreement among many raters.
\newblock \emph{Psychological Bulletin}, 76\penalty0 (5):\penalty0 378--382,
  1971.
\newblock ISSN 1939-1455(Electronic),0033-2909(Print).
\newblock \doi{10.1037/h0031619}.

\bibitem[Fu et~al.(2019)Fu, Leung, Wang, Raulli, Kallmes, Kinsman, Nelson,
  Clark, Luetmer, Kingsbury, Kent, and Liu]{fu_natural_2019}
Sunyang Fu, Lester~Y. Leung, Yanshan Wang, Anne-Olivia Raulli, David~F.
  Kallmes, Kristin~A. Kinsman, Kristoff~B. Nelson, Michael~S. Clark, Patrick~H.
  Luetmer, Paul~R. Kingsbury, David~M. Kent, and Hongfang Liu.
\newblock Natural {Language} {Processing} for the {Identification} of {Silent}
  {Brain} {Infarcts} {From} {Neuroimaging} {Reports}.
\newblock \emph{JMIR Medical Informatics}, 7\penalty0 (2):\penalty0 e12109,
  2019.
\newblock \doi{10.2196/12109}.
\newblock URL \url{https://medinform.jmir.org/2019/2/e12109/}.

\bibitem[Garg et~al.(2019)Garg, Oh, Naidech, Kording, and
  Prabhakaran]{garg_automating_2019}
Ravi Garg, Elissa Oh, Andrew Naidech, Konrad Kording, and Shyam Prabhakaran.
\newblock Automating {Ischemic} {Stroke} {Subtype} {Classification} {Using}
  {Machine} {Learning} and {Natural} {Language} {Processing}.
\newblock \emph{Journal of Stroke and Cerebrovascular Diseases}, 28\penalty0
  (7):\penalty0 2045--2051, July 2019.
\newblock ISSN 1052-3057.
\newblock \doi{10.1016/j.jstrokecerebrovasdis.2019.02.004}.
\newblock URL
  \url{http://www.sciencedirect.com/science/article/pii/S1052305719300485}.

\bibitem[Gehanno et~al.(2013)Gehanno, Rollin, and Darmoni]{gehanno_is_2013}
Jean-François Gehanno, Laetitia Rollin, and Stefan Darmoni.
\newblock Is the coverage of google scholar enough to be used alone for
  systematic reviews.
\newblock \emph{BMC Medical Informatics and Decision Making}, 13:\penalty0 7,
  2013.
\newblock \doi{10.1186/1472-6947-13-7}.
\newblock URL \url{https://www.ncbi.nlm.nih.gov/pmc/articles/PMC3544576/}.

\bibitem[Goldshtein et~al.(2020)Goldshtein, Chodick, Kochba, Gal, Webb, and
  Shibolet]{goldshtein_identification_2020}
Inbal Goldshtein, Gabriel Chodick, Ilan Kochba, Nitsan Gal, Muriel Webb, and
  Oren Shibolet.
\newblock Identification and {Characterization} of {Nonalcoholic} {Fatty}
  {Liver} {Disease}.
\newblock \emph{Clinical Gastroenterology and Hepatology}, 18\penalty0
  (8):\penalty0 1887--1889, July 2020.
\newblock ISSN 1542-3565.
\newblock \doi{10.1016/j.cgh.2019.08.007}.
\newblock URL
  \url{http://www.sciencedirect.com/science/article/pii/S1542356519308638}.

\bibitem[Gorinski et~al.(2019)Gorinski, Wu, Grover, Tobin, Talbot, Whalley,
  Sudlow, Whiteley, and Alex]{gorinski_named_2019}
Philip~John Gorinski, Honghan Wu, Claire Grover, Richard Tobin, Conn Talbot,
  Heather Whalley, Cathie Sudlow, William Whiteley, and Beatrice Alex.
\newblock Named {Entity} {Recognition} for {Electronic} {Health} {Records}: {A}
  {Comparison} of {Rule}-based and {Machine} {Learning} {Approaches}.
\newblock \emph{arXiv:1903.03985 [cs.CL]}, June 2019.
\newblock URL \url{http://arxiv.org/abs/1903.03985}.

\bibitem[Gould et~al.(2015)Gould, Tang, Liu, Lee, Zheng, Danforth, Kosco,
  Di~Fiore, and Suh]{gould_recent_2015}
Michael~K. Gould, Tania Tang, In-Lu~Amy Liu, Janet Lee, Chengyi Zheng, Kim~N.
  Danforth, Anne~E. Kosco, Jamie~L. Di~Fiore, and David~E. Suh.
\newblock Recent {Trends} in the {Identification} of {Incidental} {Pulmonary}
  {Nodules}.
\newblock \emph{American Journal of Respiratory and Critical Care Medicine},
  192\penalty0 (10):\penalty0 1208--1214, July 2015.
\newblock ISSN 1073-449X.
\newblock \doi{10.1164/rccm.201505-0990OC}.
\newblock URL
  \url{https://www.atsjournals.org/doi/full/10.1164/rccm.201505-0990OC}.

\bibitem[Grivas et~al.(2020)Grivas, Alex, Grover, {Tobin, R.}, and {Whiteley,
  W.}]{grivas_not_2020}
A.~Grivas, B.~Alex, C.~Grover, {Tobin, R.}, and {Whiteley, W.}
\newblock Not a cute stroke: {Analysis} of {Rule}- and {Neural}
  {Network}-{Based} {Information} {Extraction} {Systems} for {Brain}
  {Radiology} {Reports}.
\newblock In \emph{Proceedings of the 11th {International} {Workshop} on
  {Health} {Text} {Mining} and {Information} {Analysis}}, 2020.

\bibitem[Grundmeier et~al.(2016)Grundmeier, Masino, Casper, Dean, Bell,
  Enriquez, Deakyne, Chamberlain, and Alpern]{grundmeier_identification_2016}
Robert~W. Grundmeier, Aaron~J. Masino, T.~Charles Casper, Jonathan~M. Dean,
  Jamie Bell, Rene Enriquez, Sara Deakyne, James~M. Chamberlain, and
  Elizabeth~R. Alpern.
\newblock Identification of {Long} {Bone} {Fractures} in {Radiology} {Reports}
  {Using} {Natural} {Language} {Processing} to {Support} {Healthcare} {Quality}
  {Improvement}.
\newblock \emph{Applied Clinical Informatics}, 7\penalty0 (4):\penalty0
  1051--1068, November 2016.
\newblock ISSN 1869-0327.
\newblock \doi{10.4338/ACI-2016-08-RA-0129}.
\newblock URL \url{https://www.ncbi.nlm.nih.gov/pmc/articles/PMC5228143/}.

\bibitem[Gupta et~al.(2018)Gupta, Banerjee, and Rubin]{gupta_automatic_2018}
Anupama Gupta, Imon Banerjee, and Daniel~L. Rubin.
\newblock Automatic information extraction from unstructured mammography
  reports using distributed semantics.
\newblock \emph{Journal of Biomedical Informatics}, 78:\penalty0 78--86,
  February 2018.
\newblock ISSN 1532-0464.
\newblock \doi{10.1016/j.jbi.2017.12.016}.
\newblock URL
  \url{http://www.sciencedirect.com/science/article/pii/S1532046417302903}.

\bibitem[{Harzing A. W.}(2007)]{harzing_a_w_publish_2007}
{Harzing A. W.}
\newblock \emph{Publish or {Perish}}.
\newblock 2007.
\newblock URL \url{Available from
  https://harzing.com/resources/publish-or-perish}.

\bibitem[Hassanpour and Langlotz(2016)]{hassanpour_unsupervised_2016}
Saeed Hassanpour and Curtis~P. Langlotz.
\newblock Unsupervised {Topic} {Modeling} in a {Large} {Free} {Text}
  {Radiology} {Report} {Repository}.
\newblock \emph{Journal of Digital Imaging}, 29\penalty0 (1):\penalty0 59--62,
  February 2016.
\newblock ISSN 1618-727X.
\newblock \doi{10.1007/s10278-015-9823-3}.
\newblock URL \url{https://doi.org/10.1007/s10278-015-9823-3}.

\bibitem[Hassanpour et~al.(2017)Hassanpour, Bay, and
  Langlotz]{hassanpour_characterization_2017}
Saeed Hassanpour, Graham Bay, and Curtis~P. Langlotz.
\newblock Characterization of {Change} and {Significance} for {Clinical}
  {Findings} in {Radiology} {Reports} {Through} {Natural} {Language}
  {Processing}.
\newblock \emph{Journal of Digital Imaging}, 30\penalty0 (3):\penalty0
  314--322, June 2017.
\newblock ISSN 1618-727X.
\newblock \doi{10.1007/s10278-016-9931-8}.
\newblock URL \url{https://doi.org/10.1007/s10278-016-9931-8}.

\bibitem[Heilbrun et~al.(2019)Heilbrun, Chapman, Narasimhan, Patel, and
  Mowery]{heilbrun_feasibility_2019}
Marta~E. Heilbrun, Brian~E. Chapman, Evan Narasimhan, Neel Patel, and Danielle
  Mowery.
\newblock Feasibility of {Natural} {Language} {Processing}–{Assisted}
  {Auditing} of {Critical} {Findings} in {Chest} {Radiology}.
\newblock \emph{Journal of the American College of Radiology}, 16\penalty0 (9,
  Part B):\penalty0 1299--1304, September 2019.
\newblock ISSN 1546-1440.
\newblock \doi{10.1016/j.jacr.2019.05.038}.
\newblock URL
  \url{http://www.sciencedirect.com/science/article/pii/S1546144019306386}.

\bibitem[Hong and Zhang(2015)]{hong_investigation_2015}
Yi~Hong and Jin Zhang.
\newblock Investigation of {Terminology} {Coverage} in {Radiology} {Reporting}
  {Templates} and {Free}‐text {Reports}.
\newblock \emph{International Journal of Knowledge Content Development \&
  Technology}, 5:\penalty0 5--14, 2015.
\newblock \doi{10.5865/IJKCT.2015.5.1.005}.

\bibitem[Huhdanpaa et~al.(2018)Huhdanpaa, Tan, Rundell, Suri, Chokshi,
  Comstock, Heagerty, James, Avins, Nedeljkovic, Nerenz, Kallmes, Luetmer,
  Sherman, Organ, Griffith, Langlotz, Carrell, Hassanpour, and
  Jarvik]{huhdanpaa_using_2018}
Hannu~T. Huhdanpaa, W.~Katherine Tan, Sean~D. Rundell, Pradeep Suri, Falgun~H.
  Chokshi, Bryan~A. Comstock, Patrick~J. Heagerty, Kathryn~T. James, Andrew~L.
  Avins, Srdjan~S. Nedeljkovic, David~R. Nerenz, David~F. Kallmes, Patrick~H.
  Luetmer, Karen~J. Sherman, Nancy~L. Organ, Brent Griffith, Curtis~P.
  Langlotz, David Carrell, Saeed Hassanpour, and Jeffrey~G. Jarvik.
\newblock Using {Natural} {Language} {Processing} of {Free}-{Text} {Radiology}
  {Reports} to {Identify} {Type} 1 {Modic} {Endplate} {Changes}.
\newblock \emph{Journal of Digital Imaging}, 31\penalty0 (1):\penalty0 84--90,
  February 2018.
\newblock ISSN 1618-727X.
\newblock \doi{10.1007/s10278-017-0013-3}.
\newblock URL \url{https://doi.org/10.1007/s10278-017-0013-3}.

\bibitem[Jnawali et~al.(2019)Jnawali, Arbabshirani, Ulloa, Rao, and
  Patel]{jnawali_automatic_2019}
K.~Jnawali, M.~R. Arbabshirani, A.~E. Ulloa, N.~Rao, and A.~A. Patel.
\newblock Automatic {Classification} of {Radiological} {Report} for
  {Intracranial} {Hemorrhage}.
\newblock In \emph{2019 {IEEE} 13th {International} {Conference} on {Semantic}
  {Computing} ({ICSC})}, pages 187--190, Newport Beach, CA, USA, January 2019.
  IEEE.
\newblock \doi{10.1109/ICOSC.2019.8665578}.

\bibitem[Johnson et~al.(2015)Johnson, Baughman, and
  Ozsoyoglu]{johnson_method_2015}
E.~Johnson, W.~C. Baughman, and G.~Ozsoyoglu.
\newblock A method for imputation of semantic class in diagnostic radiology
  text.
\newblock In \emph{2015 {IEEE} {International} {Conference} on {Bioinformatics}
  and {Biomedicine} ({BIBM})}, pages 750--755, Washington, DC, November 2015.
  IEEE.
\newblock \doi{10.1109/BIBM.2015.7359780}.

\bibitem[Kang et~al.(2019)Kang, Garry, Chung, Moore, Iturrate, Swartz, Kim,
  Horwitz, and Blecker]{kang_natural_2019}
Stella~K. Kang, Kira Garry, Ryan Chung, William~H. Moore, Eduardo Iturrate,
  Jordan~L. Swartz, Danny~C. Kim, Leora~I. Horwitz, and Saul Blecker.
\newblock Natural {Language} {Processing} for {Identification} of {Incidental}
  {Pulmonary} {Nodules} in {Radiology} {Reports}.
\newblock \emph{Journal of the American College of Radiology}, 16\penalty0
  (11):\penalty0 1587--1594, November 2019.
\newblock ISSN 1546-1440.
\newblock \doi{10.1016/j.jacr.2019.04.026}.
\newblock URL
  \url{http://www.sciencedirect.com/science/article/pii/S1546144019305332}.

\bibitem[Karunakaran et~al.(2017)Karunakaran, Misra, Marshall, Mathrawala, and
  Kethireddy]{karunakaran_closing_2017}
B.~Karunakaran, D.~Misra, K.~Marshall, D.~Mathrawala, and S.~Kethireddy.
\newblock Closing the loop — {Finding} lung cancer patients using {NLP}.
\newblock In \emph{2017 {IEEE} {International} {Conference} on {Big} {Data}
  ({Big} {Data})}, pages 2452--2461, Boston, MA, December 2017. IEEE.
\newblock \doi{10.1109/BigData.2017.8258203}.

\bibitem[Kehl et~al.(2019)Kehl, Elmarakeby, Nishino, Van~Allen, Lepisto,
  Hassett, Johnson, and Schrag]{kehl_assessment_2019}
Kenneth~L. Kehl, Haitham Elmarakeby, Mizuki Nishino, Eliezer~M. Van~Allen,
  Eva~M. Lepisto, Michael~J. Hassett, Bruce~E. Johnson, and Deborah Schrag.
\newblock Assessment of {Deep} {Natural} {Language} {Processing} in
  {Ascertaining} {Oncologic} {Outcomes} {From} {Radiology} {Reports}.
\newblock \emph{JAMA Oncology}, 5\penalty0 (10):\penalty0 1421--1429, October
  2019.
\newblock ISSN 2374-2437.
\newblock \doi{10.1001/jamaoncol.2019.1800}.
\newblock URL \url{https://doi.org/10.1001/jamaoncol.2019.1800}.

\bibitem[Kim et~al.(2019)Kim, Zhu, Obeid, and Lenert]{kim_natural_2019}
Chulho Kim, Vivienne Zhu, Jihad Obeid, and Leslie Lenert.
\newblock Natural language processing and machine learning algorithm to
  identify brain {MRI} reports with acute ischemic stroke.
\newblock \emph{PLOS ONE}, 14\penalty0 (2):\penalty0 e0212778, February 2019.
\newblock ISSN 1932-6203.
\newblock \doi{10.1371/journal.pone.0212778}.
\newblock URL
  \url{https://journals.plos.org/plosone/article?id=10.1371/journal.pone.0212778}.

\bibitem[Kreimeyer et~al.(2017)Kreimeyer, Foster, Pandey, Arya, Halford, Jones,
  Forshee, Walderhaug, and Botsis]{kreimeyer_natural_2017}
Kory Kreimeyer, Matthew Foster, Abhishek Pandey, Nina Arya, Gwendolyn Halford,
  Sandra~F. Jones, Richard Forshee, Mark Walderhaug, and Taxiarchis Botsis.
\newblock Natural language processing systems for capturing and standardizing
  unstructured clinical information: {A} systematic review.
\newblock \emph{Journal of Biomedical Informatics}, 73:\penalty0 14--29, 2017.
\newblock ISSN 1532-0480.
\newblock \doi{10.1016/j.jbi.2017.07.012}.

\bibitem[Kwan et~al.(2019)Kwan, Yermak, Markell, Paul, Shojania, and
  Cram]{kwan_follow_2019}
Janice~L. Kwan, Darya Yermak, Lezlie Markell, Narinder~S. Paul, Kaveh~J.
  Shojania, and Peter Cram.
\newblock Follow {Up} of {Incidental} {High}-{Risk} {Pulmonary} {Nodules} on
  {Computed} {Tomography} {Pulmonary} {Angiography} at {Care} {Transitions}.
\newblock \emph{Journal of Hospital Medicine}, 14\penalty0 (6):\penalty0
  349--352, June 2019.
\newblock \doi{10.12788/jhm.3128}.
\newblock URL \url{https://europepmc.org/article/med/30794133}.

\bibitem[Kłos et~al.(2018)Kłos, Żyłkowski, and
  Spinczyk]{klos_automatic_2018}
Monika Kłos, Jarosław Żyłkowski, and Dominik Spinczyk.
\newblock Automatic {Classification} of {Text} {Documents} {Presenting}
  {Radiology} {Examinations}.
\newblock In Ewa Pietka, Pawel Badura, Jacek Kawa, and Wojciech Wieclawek,
  editors, \emph{Proceedings 6th {International} {Conference} {Information}
  {Technology} in {Biomedicine}({ITIB}'2018)}, Advances in {Intelligent}
  {Systems} and {Computing}, pages 495--505. Springer International Publishing,
  2018.
\newblock ISBN 978-3-319-91211-0.
\newblock \doi{10.1007/978-3-319-91211-0-43}.

\bibitem[Lacson et~al.(2015)Lacson, Harris, Brawarsky, Tosteson, Onega,
  Tosteson, Kaye, Gonzalez, Birdwell, and Haas]{lacson_evaluation_2015}
Ronilda Lacson, Kimberly Harris, Phyllis Brawarsky, Tor~D. Tosteson, Tracy
  Onega, Anna N.~A. Tosteson, Abby Kaye, Irina Gonzalez, Robyn Birdwell, and
  Jennifer~S. Haas.
\newblock Evaluation of an {Automated} {Information} {Extraction} {Tool} for
  {Imaging} {Data} {Elements} to {Populate} a {Breast} {Cancer} {Screening}
  {Registry}.
\newblock \emph{Journal of Digital Imaging}, 28\penalty0 (5):\penalty0
  567--575, October 2015.
\newblock ISSN 1618-727X.
\newblock \doi{10.1007/s10278-014-9762-4}.
\newblock URL \url{https://doi.org/10.1007/s10278-014-9762-4}.

\bibitem[Lacson et~al.(2017)Lacson, Goodrich, Harris, Brawarsky, and
  Haas]{lacson_assessing_2017}
Ronilda Lacson, Martha~E. Goodrich, Kimberly Harris, Phyllis Brawarsky, and
  Jennifer~S. Haas.
\newblock Assessing {Inaccuracies} in {Automated} {Information} {Extraction} of
  {Breast} {Imaging} {Findings}.
\newblock \emph{Journal of Digital Imaging}, 30\penalty0 (2):\penalty0
  228--233, April 2017.
\newblock ISSN 1618-727X.
\newblock \doi{10.1007/s10278-016-9927-4}.
\newblock URL \url{https://doi.org/10.1007/s10278-016-9927-4}.

\bibitem[Lafourcade and Ramadier(2017)]{lafourcade_radiological_2017}
M.~Lafourcade and Lionel Ramadier.
\newblock Radiological text simplification using a general knowledge base.
\newblock In \emph{18th {International} {Conference} on {Computational}
  {Linguistics} and {Intelligent} {Text} {Processing} ({CICLing} 2017)},
  {CICLing} 2017. Budapest, Hungary, 2017.
\newblock \doi{https://doi.org/10.1007/978-3-319-77116-8\_46}.

\bibitem[Lafourcade and Ramadier(2016)]{lafourcade_semantic_2016}
Mathieu Lafourcade and Lionel Ramadier.
\newblock Semantic {RelationExtraction} with {Semantic} {Patterns}:
  {Experiment} on {Radiology} {Report}.
\newblock In \emph{Proceedings of the {Tenth} {International} {Conference} on
  {Language} {Resources} and {Evaluation} ({LREC} 2016)}, {LREC} 2016
  {Proceedings}, Portorož, Slovenia, 2016. European Language Resources
  Association (ELRA).
\newblock ISBN 978-2-9517408-9-1.
\newblock URL \url{https://hal.archives-ouvertes.fr/hal-01382320}.

\bibitem[Landis and Koch(1977)]{landis_measurement_1977}
J.~Richard Landis and Gary~G. Koch.
\newblock The {Measurement} of {Observer} {Agreement} for {Categorical} {Data}.
\newblock \emph{Biometrics}, 33\penalty0 (1):\penalty0 159--174, 1977.
\newblock ISSN 0006-341X.
\newblock \doi{10.2307/2529310}.
\newblock URL \url{https://www.jstor.org/stable/2529310}.

\bibitem[Li and Elliot(2019)]{li_natural_2019}
Andrew~Yu Li and Nikki Elliot.
\newblock Natural language processing to identify ureteric stones in radiology
  reports.
\newblock \emph{Journal of Medical Imaging and Radiation Oncology}, 63\penalty0
  (3):\penalty0 307--310, 2019.
\newblock ISSN 1754-9485.
\newblock \doi{10.1111/1754-9485.12861}.
\newblock URL
  \url{https://onlinelibrary.wiley.com/doi/abs/10.1111/1754-9485.12861}.

\bibitem[Liu et~al.(2019)Liu, Zhu, Liu, Han, Zhang, and
  Wang]{liu_automatic_2019}
Yi~Liu, Li-Na Zhu, Qing Liu, Chao Han, Xiao-Dong Zhang, and Xiao-Ying Wang.
\newblock Automatic extraction of imaging observation and assessment categories
  from breast magnetic resonance imaging reports with natural language
  processing.
\newblock \emph{Chinese Medical Journal}, 132\penalty0 (14):\penalty0
  1673--1680, July 2019.
\newblock ISSN 0366-6999.
\newblock \doi{10.1097/CM9.0000000000000301}.
\newblock URL \url{https://www.ncbi.nlm.nih.gov/pmc/articles/PMC6759110/}.

\bibitem[Mabotuwana et~al.(2018{\natexlab{a}})Mabotuwana, Hall, Tieder, and
  Gunn]{mabotuwana_improving_2018}
Thusitha Mabotuwana, Christopher~S Hall, Joel Tieder, and Martin~L. Gunn.
\newblock Improving {Quality} of {Follow}-{Up} {Imaging} {Recommendations} in
  {Radiology}.
\newblock \emph{AMIA Annual Symposium Proceedings}, 2017:\penalty0 1196--1204,
  April 2018{\natexlab{a}}.
\newblock ISSN 1942-597X.
\newblock URL \url{https://www.ncbi.nlm.nih.gov/pmc/articles/PMC5977608/}.

\bibitem[Mabotuwana et~al.(2018{\natexlab{b}})Mabotuwana, Hombal, Dalal, Hall,
  and Gunn]{mabotuwana_determining_2018}
Thusitha Mabotuwana, Vadiraj Hombal, Sandeep Dalal, Christopher~S. Hall, and
  Martin Gunn.
\newblock Determining {Adherence} to {Follow}-up {Imaging} {Recommendations}.
\newblock \emph{Journal of the American College of Radiology}, 15\penalty0 (3,
  Part A):\penalty0 422--428, March 2018{\natexlab{b}}.
\newblock ISSN 1546-1440.
\newblock \doi{10.1016/j.jacr.2017.11.022}.
\newblock URL
  \url{http://www.sciencedirect.com/science/article/pii/S1546144017314758}.

\bibitem[Mahan et~al.(2019)Mahan, Rafter, Casey, Engelking, Abdallah, Truwit,
  Oswood, and Samadani]{mahan_tbiextractor_2019}
Margaret Mahan, Daniel Rafter, Hannah Casey, Marta Engelking, Tessneem
  Abdallah, Charles Truwit, Mark Oswood, and Uzma Samadani.
\newblock {tbiExtractor}: {A} framework for extracting traumatic brain injury
  common data elements from radiology reports.
\newblock \emph{bioRxiv 585331}, 2019.
\newblock \doi{10.1101/585331}.
\newblock URL \url{https://www.biorxiv.org/content/10.1101/585331v1}.

\bibitem[Maros et~al.(2018)Maros, Wenz, Förster, Froelich, Groden, Sommer,
  Schönberg, Henzler, and Wenz]{maros_objective_2018}
Máté~E. Maros, Ralf Wenz, Alex Förster, Matthias~F. Froelich, Christoph
  Groden, Wieland~H. Sommer, Stefan~O. Schönberg, Thomas Henzler, and Holger
  Wenz.
\newblock Objective {Comparison} {Using} {Guideline}-based {Query} of
  {Conventional} {Radiological} {Reports} and {Structured} {Reports}.
\newblock \emph{In Vivo}, 32\penalty0 (4):\penalty0 843--849, January 2018.
\newblock ISSN 0258-851X, 1791-7549.
\newblock \doi{10.21873/invivo.11318}.
\newblock URL \url{http://iv.iiarjournals.org/content/32/4/843}.

\bibitem[Masino et~al.(2016)Masino, Grundmeier, Pennington, Germiller, and
  Crenshaw]{masino_temporal_2016}
Aaron~J. Masino, Robert~W. Grundmeier, Jeffrey~W. Pennington, John~A.
  Germiller, and E.~Bryan Crenshaw.
\newblock Temporal bone radiology report classification using open source
  machine learning and natural langue processing libraries.
\newblock \emph{BMC Medical Informatics and Decision Making}, 16\penalty0
  (1):\penalty0 65, June 2016.
\newblock ISSN 1472-6947.
\newblock \doi{10.1186/s12911-016-0306-3}.
\newblock URL \url{https://doi.org/10.1186/s12911-016-0306-3}.

\bibitem[Meystre et~al.(2017)Meystre, Gouripeddi, Tieder, Simmons, Srivastava,
  and Shah]{meystre_enhancing_2017}
Stephane Meystre, Ramkiran Gouripeddi, Joel Tieder, Jeffrey Simmons, Rajendu
  Srivastava, and Samir Shah.
\newblock Enhancing {Comparative} {Effectiveness} {Research} {With} {Automated}
  {Pediatric} {Pneumonia} {Detection} in a {Multi}-{Institutional} {Clinical}
  {Repository}: {A} {PHIS}+ {Pilot} {Study}.
\newblock \emph{Journal of Medical Internet Research}, 19\penalty0
  (5):\penalty0 e162, 2017.
\newblock \doi{10.2196/jmir.6887}.
\newblock URL \url{https://www.jmir.org/2017/5/e162/}.

\bibitem[Miao et~al.(2018)Miao, Xu, Wu, Xie, Wang, Jing, Zhang, Zhang, Yang,
  Zhang, Shan, Wang, Xu, Wang, and Liu]{miao_extraction_2018}
Shumei Miao, Tingyu Xu, Yonghui Wu, Hui Xie, Jingqi Wang, Shenqi Jing, Yaoyun
  Zhang, Xiaoliang Zhang, Yinshuang Yang, Xin Zhang, Tao Shan, Li~Wang, Hua Xu,
  Shui Wang, and Yun Liu.
\newblock Extraction of {BI}-{RADS} findings from breast ultrasound reports in
  {Chinese} using deep learning approaches.
\newblock \emph{International Journal of Medical Informatics}, 119:\penalty0
  17--21, November 2018.
\newblock ISSN 1386-5056.
\newblock \doi{10.1016/j.ijmedinf.2018.08.009}.
\newblock URL
  \url{http://www.sciencedirect.com/science/article/pii/S1386505618309225}.

\bibitem[Mikolov et~al.(2013)Mikolov, Chen, Corrado, and
  Dean]{mikolov_efficient_2013}
Tomas Mikolov, Kai Chen, Greg Corrado, and Jeffrey Dean.
\newblock \emph{Efficient {Estimation} of {Word} {Representations} in {Vector}
  {Space}}.
\newblock 2013.
\newblock URL \url{http://arxiv.org/abs/1301.3781}.

\bibitem[Mikolov et~al.(2018)Mikolov, Grave, Bojanowski, Puhrsch, and
  Joulin]{mikolov_advances_2018}
Tomas Mikolov, Edouard Grave, Piotr Bojanowski, Christian Puhrsch, and Armand
  Joulin.
\newblock Advances in {Pre}-{Training} {Distributed} {Word} {Representations}.
\newblock In \emph{Proceedings of the {International} {Conference} on
  {Language} {Resources} and {Evaluation} ({LREC} 2018)}, 2018.

\bibitem[Minn et~al.(2015)Minn, Zandieh, and Filice]{minn_improving_2015}
Matthew~J. Minn, Arash~R. Zandieh, and Ross~W. Filice.
\newblock Improving {Radiology} {Report} {Quality} by {Rapidly} {Notifying}
  {Radiologist} of {Report} {Errors}.
\newblock \emph{Journal of Digital Imaging}, 28\penalty0 (4):\penalty0
  492--498, August 2015.
\newblock ISSN 1618-727X.
\newblock \doi{10.1007/s10278-015-9781-9}.
\newblock URL \url{https://doi.org/10.1007/s10278-015-9781-9}.

\bibitem[Moher et~al.(2015)Moher, Shamseer, Clarke, Ghersi, Liberati,
  Petticrew, Shekelle, and Stewart]{moher_preferred_2015}
David Moher, Larissa Shamseer, Mike Clarke, Davina Ghersi, Alessandro Liberati,
  Mark Petticrew, Paul Shekelle, and Lesley~A Stewart.
\newblock Preferred reporting items for systematic review and meta-analysis
  protocols ({PRISMA}-{P}) 2015 statement.
\newblock \emph{Systematic Reviews}, 4\penalty0 (1):\penalty0 1, December 2015.
\newblock ISSN 2046-4053.
\newblock \doi{10.1186/2046-4053-4-1}.
\newblock URL
  \url{https://systematicreviewsjournal.biomedcentral.com/articles/10.1186/2046-4053-4-1}.

\bibitem[Mujjiga et~al.(2019)Mujjiga, Krishna, Chakravarthi, and
  J]{mujjiga_identifying_2019}
Srikanth Mujjiga, Vamsi Krishna, Kalyan Chakravarthi, and Vijayananda J.
\newblock Identifying {Semantics} in {Clinical} {Reports} {Using} {Neural}
  {Machine} {Translation}.
\newblock \emph{Proceedings of the AAAI Conference on Artificial Intelligence},
  33\penalty0 (01):\penalty0 9552--9557, July 2019.
\newblock ISSN 2374-3468.
\newblock \doi{10.1609/aaai.v33i01.33019552}.
\newblock URL \url{https://www.aaai.org/ojs/index.php/AAAI/article/view/5015}.

\bibitem[{National Library of
  Medicine}(2021{\natexlab{a}})]{national_library_of_medicine_snomed_2021}
{National Library of Medicine}.
\newblock \emph{{SNOMED} {CT}}.
\newblock 2021{\natexlab{a}}.
\newblock URL \url{@miscumls, author = National Library of Medicine, title =
  SNOMED CT, url = https://www.nlm.nih.gov/healthit/snomedct/index.html, year =
  2021, urldate=Accessed 07 Feb 2021}.

\bibitem[{National Library of
  Medicine}(2021{\natexlab{b}})]{national_library_of_medicine_unified_2021}
{National Library of Medicine}.
\newblock \emph{Unified {Medical} {Language} {System}}.
\newblock 2021{\natexlab{b}}.
\newblock URL \url{https://www.nlm.nih.gov/research/umls/index.html}.

\bibitem[Noorbakhsh-Sabet et~al.(2018)Noorbakhsh-Sabet, Tsivgoulis, Shahjouei,
  Hu, Goyal, Alexandrov, and Zand]{noorbakhsh-sabet_racial_2018}
Nariman Noorbakhsh-Sabet, Georgios Tsivgoulis, Shima Shahjouei, Yirui Hu, Nitin
  Goyal, Andrei~V. Alexandrov, and Ramin Zand.
\newblock Racial {Difference} in {Cerebral} {Microbleed} {Burden} {Among} a
  {Patient} {Population} in the {Mid}-{South} {United} {States}.
\newblock \emph{Journal of Stroke and Cerebrovascular Diseases}, 27\penalty0
  (10):\penalty0 2657--2661, October 2018.
\newblock ISSN 1052-3057.
\newblock \doi{10.1016/j.jstrokecerebrovasdis.2018.05.031}.
\newblock URL
  \url{http://www.sciencedirect.com/science/article/pii/S1052305718302714}.

\bibitem[Oberkampf et~al.(2016)Oberkampf, Zillner, Overton, Bauer, Cavallaro,
  Uder, and Hammon]{oberkampf_semantic_2016}
Heiner Oberkampf, Sonja Zillner, James~A. Overton, Bernhard Bauer, Alexander
  Cavallaro, Michael Uder, and Matthias Hammon.
\newblock Semantic representation of reported measurements in radiology.
\newblock \emph{BMC Medical Informatics and Decision Making}, 16\penalty0
  (1):\penalty0 5, January 2016.
\newblock ISSN 1472-6947.
\newblock \doi{10.1186/s12911-016-0248-9}.
\newblock URL \url{https://doi.org/10.1186/s12911-016-0248-9}.

\bibitem[Ong et~al.(2020)Ong, Orfanoudaki, Zhang, Caprasse, Hutch, Ma, Fard,
  Balogun, Miller, Minnig, Saglam, Prescott, Greer, Smirnakis, and
  Bertsimas]{ong_machine_2020}
Charlene~Jennifer Ong, Agni Orfanoudaki, Rebecca Zhang, Francois Pierre~M.
  Caprasse, Meghan Hutch, Liang Ma, Darian Fard, Oluwafemi Balogun, Matthew~I.
  Miller, Margaret Minnig, Hanife Saglam, Brenton Prescott, David~M. Greer,
  Stelios Smirnakis, and Dimitris Bertsimas.
\newblock Machine learning and natural language processing methods to identify
  ischemic stroke, acuity and location from radiology reports.
\newblock \emph{PLOS ONE}, 15\penalty0 (6):\penalty0 e0234908, June 2020.
\newblock ISSN 1932-6203.
\newblock \doi{10.1371/journal.pone.0234908}.
\newblock URL
  \url{https://journals.plos.org/plosone/article?id=10.1371/journal.pone.0234908}.

\bibitem[Patel et~al.(2017)Patel, Puppala, Ogunti, Ensor, He, Shewale, Ankerst,
  Kaklamani, Rodriguez, Wong, and Chang]{patel_correlating_2017}
Tejal~A. Patel, Mamta Puppala, Richard~O. Ogunti, Joe~E. Ensor, Tiancheng He,
  Jitesh~B. Shewale, Donna~P. Ankerst, Virginia~G. Kaklamani, Angel~A.
  Rodriguez, Stephen T.~C. Wong, and Jenny~C. Chang.
\newblock Correlating mammographic and pathologic findings in clinical decision
  support using natural language processing and data mining methods.
\newblock \emph{Cancer}, 123\penalty0 (1):\penalty0 114--121, January 2017.
\newblock ISSN 1097-0142.
\newblock \doi{10.1002/cncr.30245}.

\bibitem[Peng et~al.(2019)Peng, Yan, Sandfort, Summers, and
  Lu]{peng_self-attention_2019}
Y.~Peng, K.~Yan, V.~Sandfort, R.~M. Summers, and Z.~Lu.
\newblock A self-attention based deep learning method for lesion attribute
  detection from {CT} reports.
\newblock In \emph{2019 {IEEE} {International} {Conference} on {Healthcare}
  {Informatics} ({ICHI})}, pages 1--5, Xi'an, China, June 2019. IEEE Computer
  Society.
\newblock \doi{10.1109/ICHI.2019.8904668}.

\bibitem[Pennington et~al.(2014)Pennington, Socher, and
  Manning]{pennington_glove_2014}
Jeffrey Pennington, Richard Socher, and Christopher~D Manning.
\newblock Glove: {Global} vectors for word representation.
\newblock In \emph{Proceedings of the 2014 conference on empirical methods in
  natural language processing ({EMNLP})}, pages 1532--1543, 2014.

\bibitem[Percha et~al.(2018)Percha, Zhang, Bozkurt, Rubin, Altman, and
  Langlotz]{percha_expanding_2018}
Bethany Percha, Yuhao Zhang, Selen Bozkurt, Daniel Rubin, Russ~B. Altman, and
  Curtis~P. Langlotz.
\newblock Expanding a radiology lexicon using contextual patterns in radiology
  reports.
\newblock \emph{Journal of the American Medical Informatics Association},
  25\penalty0 (6):\penalty0 679--685, June 2018.
\newblock \doi{10.1093/jamia/ocx152}.
\newblock URL \url{https://academic.oup.com/jamia/article/25/6/679/4797401}.

\bibitem[Peters et~al.(2018)Peters, Neumann, Iyyer, Gardner, Clark, Lee, and
  Zettlemoyer]{peters_deep_2018}
Matthew~E. Peters, Mark Neumann, Mohit Iyyer, Matt Gardner, Christopher Clark,
  Kenton Lee, and Luke Zettlemoyer.
\newblock Deep contextualized word representations.
\newblock \emph{CoRR}, abs/1802.05365, 2018.
\newblock URL \url{http://arxiv.org/abs/1802.05365}.
\newblock \_eprint: 1802.05365.

\bibitem[Pons et~al.(2016)Pons, Braun, Hunink, and Kors]{pons_natural_2016}
Ewoud Pons, Loes M.~M. Braun, M.~G.~Myriam Hunink, and Jan~A. Kors.
\newblock Natural {Language} {Processing} in {Radiology}: {A} {Systematic}
  {Review}.
\newblock \emph{Radiology}, 279\penalty0 (2):\penalty0 329--343, April 2016.
\newblock ISSN 0033-8419.
\newblock \doi{10.1148/radiol.16142770}.
\newblock URL \url{https://pubs.rsna.org/doi/10.1148/radiol.16142770}.

\bibitem[Pruitt et~al.(2019)Pruitt, Naidech, Van~Ornam, Borczuk, and
  Thompson]{pruitt_natural_2019}
Peter Pruitt, Andrew Naidech, Jonathan Van~Ornam, Pierre Borczuk, and William
  Thompson.
\newblock A natural language processing algorithm to extract characteristics of
  subdural hematoma from head {CT} reports.
\newblock \emph{Emergency Radiology}, 26\penalty0 (3):\penalty0 301--306, June
  2019.
\newblock ISSN 1438-1435.
\newblock \doi{10.1007/s10140-019-01673-4}.
\newblock URL \url{https://doi.org/10.1007/s10140-019-01673-4}.

\bibitem[Qenam et~al.(2017)Qenam, Kim, Carroll, and Hogarth]{qenam_text_2017}
Basel Qenam, Tae~Youn Kim, Mark~J. Carroll, and Michael Hogarth.
\newblock Text {Simplification} {Using} {Consumer} {Health} {Vocabulary} to
  {Generate} {Patient}-{Centered} {Radiology} {Reporting}: {Translation} and
  {Evaluation}.
\newblock \emph{Journal of Medical Internet Research}, 19\penalty0
  (12):\penalty0 e417, 2017.
\newblock \doi{10.2196/jmir.8536}.
\newblock URL \url{https://www.jmir.org/2017/12/e417/}.

\bibitem[Ratner et~al.(2018)Ratner, Hancock, Dunnmon, Goldman, and
  Ré]{ratner_snorkel_2018}
Alex Ratner, Braden Hancock, Jared Dunnmon, Roger Goldman, and Christopher Ré.
\newblock Snorkel {MeTaL}: {Weak} {Supervision} for {Multi}-{Task} {Learning}.
\newblock In \emph{Proceedings of the {Second} {Workshop} on {Data}
  {Management} for {End}-{To}-{End} {Machine} {Learning}}, volume~3 of
  \emph{{DEEM}'18}, pages 1--4, Houston, TX, USA, 2018. ACM.
\newblock ISBN 978-1-4503-5828-6.
\newblock \doi{10.1145/3209889.3209898}.
\newblock URL \url{https://doi.org/10.1145/3209889.3209898}.

\bibitem[Redman et~al.(2017)Redman, Natarajan, Hou, Wang, Hanif, Feng, Kramer,
  Desiderio, Xu, El-Serag, and Kanwal]{redman_accurate_2017}
Joseph~S. Redman, Yamini Natarajan, Jason~K. Hou, Jingqi Wang, Muzammil Hanif,
  Hua Feng, Jennifer~R. Kramer, Roxanne Desiderio, Hua Xu, Hashem~B. El-Serag,
  and Fasiha Kanwal.
\newblock Accurate {Identification} of {Fatty} {Liver} {Disease} in {Data}
  {Warehouse} {Utilizing} {Natural} {Language} {Processing}.
\newblock \emph{Digestive Diseases and Sciences}, 62\penalty0 (10):\penalty0
  2713--2718, October 2017.
\newblock ISSN 1573-2568.
\newblock \doi{10.1007/s10620-017-4721-9}.
\newblock URL \url{https://doi.org/10.1007/s10620-017-4721-9}.

\bibitem[{RSNA}(2021)]{rsna_radlex_2021}
{RSNA}.
\newblock \emph{{RadLex}}.
\newblock 2021.
\newblock URL \url{http://radlex.org/}.

\bibitem[Sada et~al.(2016)Sada, Hou, Richardson, El-Serag, and
  Davila]{sada_validation_2016}
Yvonne Sada, Jason Hou, Peter Richardson, Hashem El-Serag, and Jessica Davila.
\newblock Validation of {Case} {Finding} {Algorithms} for {Hepatocellular}
  {Cancer} from {Administrative} {Data} and {Electronic} {Health} {Records}
  using {Natural} {Language} {Processing}.
\newblock \emph{Medical care}, 54\penalty0 (2):\penalty0 e9--e14, February
  2016.
\newblock ISSN 0025-7079.
\newblock \doi{10.1097/MLR.0b013e3182a30373}.
\newblock URL \url{https://www.ncbi.nlm.nih.gov/pmc/articles/PMC3875602/}.

\bibitem[Sevenster et~al.(2015{\natexlab{a}})Sevenster, Buurman, Liu, Peters,
  and Chang]{sevenster_natural_2015-1}
M.~Sevenster, J.~Buurman, P.~Liu, J.~F. Peters, and P.~J. Chang.
\newblock Natural {Language} {Processing} {Techniques} for {Extracting} and
  {Categorizing} {Finding} {Measurements} in {Narrative} {Radiology} {Reports}.
\newblock \emph{Applied Clinical Informatics}, 06\penalty0 (3):\penalty0
  600--610, 2015{\natexlab{a}}.
\newblock ISSN 1869-0327.
\newblock \doi{10.4338/ACI-2014-11-RA-0110}.
\newblock URL
  \url{http://www.thieme-connect.de/DOI/DOI?10.4338/ACI-2014-11-RA-0110}.

\bibitem[Sevenster et~al.(2015{\natexlab{b}})Sevenster, Bozeman, Cowhy, and
  Trost]{sevenster_natural_2015}
Merlijn Sevenster, Jeffrey Bozeman, Andrea Cowhy, and William Trost.
\newblock A natural language processing pipeline for pairing measurements
  uniquely across free-text {CT} reports.
\newblock \emph{Journal of Biomedical Informatics}, 53:\penalty0 36--48,
  February 2015{\natexlab{b}}.
\newblock ISSN 1532-0464.
\newblock \doi{10.1016/j.jbi.2014.08.015}.
\newblock URL
  \url{http://www.sciencedirect.com/science/article/pii/S1532046414001968}.

\bibitem[Shelmerdine et~al.(2019)Shelmerdine, Singh, Norman, Jones, Sebire, and
  Arthurs]{shelmerdine_automated_2019}
S.~C. Shelmerdine, M.~Singh, W.~Norman, R.~Jones, N.~J. Sebire, and O.~J.
  Arthurs.
\newblock Automated data extraction and report analysis in computer-aided
  radiology audit: practice implications from post-mortem paediatric imaging.
\newblock \emph{Clinical Radiology}, 74\penalty0 (9):\penalty0
  733.e11--733.e18, September 2019.
\newblock ISSN 0009-9260.
\newblock \doi{10.1016/j.crad.2019.04.021}.
\newblock URL
  \url{http://www.sciencedirect.com/science/article/pii/S0009926019302181}.

\bibitem[Shickel et~al.(2018)Shickel, Tighe, Bihorac, and
  Rashidi]{shickel_deep_2018}
B.~Shickel, P.~J. Tighe, A.~Bihorac, and P.~Rashidi.
\newblock Deep {EHR}: {A} {Survey} of {Recent} {Advances} in {Deep} {Learning}
  {Techniques} for {Electronic} {Health} {Record} ({EHR}) {Analysis}.
\newblock \emph{IEEE Journal of Biomedical and Health Informatics}, 22\penalty0
  (5):\penalty0 1589--1604, September 2018.
\newblock ISSN 2168-2208.
\newblock \doi{10.1109/JBHI.2017.2767063}.

\bibitem[Shin et~al.(2017)Shin, Chokshi, Lee, and
  Choi]{shin_classification_2017}
B.~Shin, F.~H. Chokshi, T.~Lee, and J.~D. Choi.
\newblock Classification of radiology reports using neural attention models.
\newblock In \emph{2017 {International} {Joint} {Conference} on {Neural}
  {Networks} ({IJCNN})}, pages 4363--4370, Anchorage, AK, May 2017. IEEE.
\newblock \doi{10.1109/IJCNN.2017.7966408}.

\bibitem[Short et~al.(2019)Short, Bralich, Bogaty, and
  Befera]{short_comprehensive_2019}
Ryan~G. Short, John Bralich, Dave Bogaty, and Nicholas~T. Befera.
\newblock Comprehensive {Word}-{Level} {Classification} of {Screening}
  {Mammography} {Reports} {Using} a {Neural} {Network} {Sequence} {Labeling}
  {Approach}.
\newblock \emph{Journal of Digital Imaging}, 32\penalty0 (5):\penalty0
  685--692, October 2019.
\newblock ISSN 1618-727X.
\newblock \doi{10.1007/s10278-018-0141-4}.
\newblock URL \url{https://doi.org/10.1007/s10278-018-0141-4}.

\bibitem[Smit et~al.(2020{\natexlab{a}})Smit, Jain, Rajpurkar, Pareek, Ng, and
  Lungren]{smit_combining_2020}
Akshay Smit, Saahil Jain, Pranav Rajpurkar, Anuj Pareek, Andrew Ng, and Matthew
  Lungren.
\newblock Combining {Automatic} {Labelers} and {Expert} {Annotations} for
  {Accurate} {Radiology} {Report} {Labeling} {Using} {BERT}.
\newblock In \emph{Proceedings of the 2020 {Conference} on {Empirical}
  {Methods} in {Natural} {Language} {Processing} ({EMNLP})}, pages 1500--1519,
  Online, November 2020{\natexlab{a}}. Association for Computational
  Linguistics.
\newblock \doi{10.18653/v1/2020.emnlp-main.117}.
\newblock URL \url{https://www.aclweb.org/anthology/2020.emnlp-main.117}.

\bibitem[Smit et~al.(2020{\natexlab{b}})Smit, Jain, Rajpurkar, Pareek, Ng, and
  Lungren]{smit_chexbert_2020}
Akshay Smit, Saahil Jain, Pranav Rajpurkar, Anuj Pareek, Andrew~Y. Ng, and
  Matthew~P. Lungren.
\newblock {CheXbert}: {Combining} {Automatic} {Labelers} and {Expert}
  {Annotations} for {Accurate} {Radiology} {Report} {Labeling} {Using} {BERT}.
\newblock \emph{CoRR}, abs/2004.09167, 2020{\natexlab{b}}.
\newblock URL \url{https://arxiv.org/abs/2004.09167}.
\newblock \_eprint: 2004.09167.

\bibitem[Sorin et~al.(2020)Sorin, Barash, Konen, and Klang]{sorin_deep_2020}
V.~Sorin, Y.~Barash, E.~Konen, and E.~Klang.
\newblock Deep {Learning} for {Natural} {Language} {Processing} in
  {Radiology}-{Fundamentals} and a {Systematic} {Review}.
\newblock \emph{Journal of the American College of Radiology : JACR},
  17\penalty0 (5):\penalty0 639--648, 2020.
\newblock \doi{10.1016/j.jacr.2019.12.026}.

\bibitem[Spasic and Nenadic(2020)]{spasic_clinical_2020}
Irena Spasic and Goran Nenadic.
\newblock Clinical {Text} {Data} in {Machine} {Learning}: {Systematic}
  {Review}.
\newblock \emph{JMIR medical informatics}, 8\penalty0 (3):\penalty0 e17984,
  March 2020.
\newblock ISSN 2291-9694.
\newblock \doi{10.2196/17984}.

\bibitem[Steinkamp et~al.(2019)Steinkamp, Chambers, Lalevic, Zafar, and
  Cook]{steinkamp_toward_2019}
Jackson~M. Steinkamp, Charles Chambers, Darco Lalevic, Hanna~M. Zafar, and
  Tessa~S. Cook.
\newblock Toward {Complete} {Structured} {Information} {Extraction} from
  {Radiology} {Reports} {Using} {Machine} {Learning}.
\newblock \emph{Journal of Digital Imaging}, 32\penalty0 (4):\penalty0
  554--564, August 2019.
\newblock ISSN 1618-727X.
\newblock \doi{10.1007/s10278-019-00234-y}.
\newblock URL \url{https://doi.org/10.1007/s10278-019-00234-y}.

\bibitem[Tahmasebi et~al.(2019)Tahmasebi, Zhu, Mankovich, Prinsen, Klassen,
  Pilato, van Ommering, Patel, Gunn, and Chang]{tahmasebi_automatic_2019}
Amir~M. Tahmasebi, Henghui Zhu, Gabriel Mankovich, Peter Prinsen, Prescott
  Klassen, Sam Pilato, Rob van Ommering, Pritesh Patel, Martin~L. Gunn, and
  Paul Chang.
\newblock Automatic {Normalization} of {Anatomical} {Phrases} in {Radiology}
  {Reports} {Using} {Unsupervised} {Learning}.
\newblock \emph{Journal of Digital Imaging}, 32\penalty0 (1):\penalty0 6--18,
  February 2019.
\newblock ISSN 1618-727X.
\newblock \doi{10.1007/s10278-018-0116-5}.
\newblock URL \url{https://doi.org/10.1007/s10278-018-0116-5}.

\bibitem[Tan and Heagerty(2019)]{tan_surrogate-guided_2019}
W.~Katherine Tan and Patrick~J. Heagerty.
\newblock Surrogate-guided sampling designs for classification of rare outcomes
  from electronic medical records data.
\newblock \emph{arXiv:1904.00412 [stat.ME]}, March 2019.
\newblock URL \url{http://arxiv.org/abs/1904.00412}.

\bibitem[Tan et~al.(2018)Tan, Hassanpour, Heagerty, Rundell, Suri, Huhdanpaa,
  James, Carrell, Langlotz, Organ, Meier, Sherman, Kallmes, Luetmer, Griffith,
  Nerenz, and Jarvik]{tan_comparison_2018}
W.~Katherine Tan, Saeed Hassanpour, Patrick~J. Heagerty, Sean~D. Rundell,
  Pradeep Suri, Hannu~T. Huhdanpaa, Kathryn James, David~S. Carrell, Curtis~P.
  Langlotz, Nancy~L. Organ, Eric~N. Meier, Karen~J. Sherman, David~F. Kallmes,
  Patrick~H. Luetmer, Brent Griffith, David~R. Nerenz, and Jeffrey~G. Jarvik.
\newblock Comparison of {Natural} {Language} {Processing} {Rules}-based and
  {Machine}-learning {Systems} to {Identify} {Lumbar} {Spine} {Imaging}
  {Findings} {Related} to {Low} {Back} {Pain}.
\newblock \emph{Academic Radiology}, 25\penalty0 (11):\penalty0 1422--1432,
  November 2018.
\newblock ISSN 1076-6332.
\newblock \doi{10.1016/j.acra.2018.03.008}.
\newblock URL
  \url{http://www.sciencedirect.com/science/article/pii/S1076633218301211}.

\bibitem[Trivedi et~al.(2019)Trivedi, Hong, Dadashzadeh, Handzel, Hochheiser,
  and Visweswaran]{trivedi_identifying_2019}
Gaurav Trivedi, Charmgil Hong, Esmaeel~R. Dadashzadeh, Robert~M. Handzel, Harry
  Hochheiser, and Shyam Visweswaran.
\newblock Identifying incidental findings from radiology reports of trauma
  patients: {An} evaluation of automated feature representation methods.
\newblock \emph{International Journal of Medical Informatics}, 129:\penalty0
  81--87, September 2019.
\newblock ISSN 1386-5056.
\newblock \doi{10.1016/j.ijmedinf.2019.05.021}.
\newblock URL
  \url{http://www.sciencedirect.com/science/article/pii/S138650561831061X}.

\bibitem[Trivedi et~al.(2018)Trivedi, Mesterhazy, Laguna, Vu, and
  Sohn]{trivedi_automatic_2018}
Hari Trivedi, Joseph Mesterhazy, Benjamin Laguna, Thienkhai Vu, and Jae~Ho
  Sohn.
\newblock Automatic {Determination} of the {Need} for {Intravenous} {Contrast}
  in {Musculoskeletal} {MRI} {Examinations} {Using} {IBM} {Watson}’s
  {Natural} {Language} {Processing} {Algorithm}.
\newblock \emph{Journal of Digital Imaging}, 31\penalty0 (2):\penalty0
  245--251, April 2018.
\newblock ISSN 1618-727X.
\newblock \doi{10.1007/s10278-017-0021-3}.
\newblock URL \url{https://doi.org/10.1007/s10278-017-0021-3}.

\bibitem[Van~Haren et~al.(2019)Van~Haren, Correa, Sepesi, Rice, Hofstetter,
  Mehran, Vaporciyan, Walsh, Roth, Swisher, and
  Antonoff]{van_haren_ground_2019}
Robert~M. Van~Haren, Arlene~M. Correa, Boris Sepesi, David~C. Rice, Wayne~L.
  Hofstetter, Reza~J. Mehran, Ara~A. Vaporciyan, Garrett~L. Walsh, Jack~A.
  Roth, Stephen~G. Swisher, and Mara~B. Antonoff.
\newblock Ground {Glass} {Lesions} on {Chest} {Imaging}: {Evaluation} of
  {Reported} {Incidence} in {Cancer} {Patients} {Using} {Natural} {Language}
  {Processing}.
\newblock \emph{The Annals of Thoracic Surgery}, 107\penalty0 (3):\penalty0
  936--940, March 2019.
\newblock ISSN 0003-4975.
\newblock \doi{10.1016/j.athoracsur.2018.09.016}.
\newblock URL
  \url{http://www.sciencedirect.com/science/article/pii/S0003497518315315}.

\bibitem[Wheater et~al.(2019)Wheater, Mair, Sudlow, Alex, Grover, and
  Whiteley]{wheater_validated_2019}
Emily Wheater, Grant Mair, Cathie Sudlow, Beatrice Alex, Claire Grover, and
  William Whiteley.
\newblock A validated natural language processing algorithm for brain imaging
  phenotypes from radiology reports in {UK} electronic health records.
\newblock \emph{BMC Medical Informatics and Decision Making}, 19\penalty0
  (1):\penalty0 184, September 2019.
\newblock ISSN 1472-6947.
\newblock \doi{10.1186/s12911-019-0908-7}.
\newblock URL \url{https://doi.org/10.1186/s12911-019-0908-7}.

\bibitem[Wohlin(2014)]{wohlin_guidelines_2014}
Claes Wohlin.
\newblock Guidelines for {Snowballing} in {Systematic} {Literature} {Studies}
  and a {Replication} in {Software} {Engineering}.
\newblock In \emph{Proceedings of the 18th {International} {Conference} on
  {Evaluation} and {Assessment} in {Software} {Engineering}}, {EASE} '14, New
  York, NY, USA, 2014. Association for Computing Machinery.
\newblock ISBN 978-1-4503-2476-2.
\newblock \doi{10.1145/2601248.2601268}.
\newblock URL \url{https://doi.org/10.1145/2601248.2601268}.
\newblock event-place: London, England, United Kingdom.

\bibitem[Wood et~al.(2020)Wood, Lynch, Kafiabadi, Guilhem, Busaidi, Montvila,
  Varsavsky, Siddiqui, Gadapa, Townend, Kiik, Patel, Barker, Ourselin, Cole,
  and Booth]{wood_automated_2020}
David~A. Wood, Jeremy Lynch, Sina Kafiabadi, Emily Guilhem, Aisha~Al Busaidi,
  Antanas Montvila, Thomas Varsavsky, Juveria Siddiqui, Naveen Gadapa, Matthew
  Townend, Martin Kiik, Keena Patel, Gareth Barker, Sebastian Ourselin,
  James~H. Cole, and Thomas~C. Booth.
\newblock Automated {Labelling} using an {Attention} model for {Radiology}
  reports of {MRI} scans ({ALARM}).
\newblock \emph{arXiv:2002.06588 [cs.CV]}, 2020.
\newblock URL \url{http://arxiv.org/abs/2002.06588}.

\bibitem[Wu et~al.(2020)Wu, Roberts, Datta, Du, Ji, Si, Soni, Wang, Wei, Xiang,
  Zhao, and Xu]{wu_deep_2020}
Stephen Wu, Kirk Roberts, Surabhi Datta, Jingcheng Du, Zongcheng Ji, Yuqi Si,
  Sarvesh Soni, Qiong Wang, Qiang Wei, Yang Xiang, Bo~Zhao, and Hua Xu.
\newblock Deep learning in clinical natural language processing: a methodical
  review.
\newblock \emph{Journal of the American Medical Informatics Association:
  JAMIA}, 27\penalty0 (3):\penalty0 457--470, 2020.
\newblock ISSN 1527-974X.
\newblock \doi{10.1093/jamia/ocz200}.

\bibitem[Xie et~al.(2019)Xie, Yang, Wang, Li, Huang, Zheng, Shu, and
  Ling]{xie_introducing_2019}
Zhe Xie, Yuanyuan Yang, Mingqing Wang, Ming Li, Haozhe Huang, Dezhong Zheng,
  Rong Shu, and Tonghui Ling.
\newblock Introducing {Information} {Extraction} to {Radiology} {Information}
  {Systems} to {Improve} the {Efficiency} on {Reading} {Reports}.
\newblock \emph{Methods of Information in Medicine}, 58\penalty0
  (2-03):\penalty0 94--106, 2019.
\newblock ISSN 2511-705X.
\newblock \doi{10.1055/s-0039-1694992}.

\bibitem[Yadav et~al.(2016)Yadav, Sarioglu, Choi, Cartwright, Hinds, and
  Chamberlain]{yadav_automated_2016}
Kabir Yadav, Efsun Sarioglu, Hyeong-Ah Choi, Walter~B. Cartwright, Pamela~S.
  Hinds, and James~M. Chamberlain.
\newblock Automated {Outcome} {Classification} of {Computed} {Tomography}
  {Imaging} {Reports} for {Pediatric} {Traumatic} {Brain} {Injury}.
\newblock \emph{Academic Emergency Medicine}, 23\penalty0 (2):\penalty0
  171--178, 2016.
\newblock ISSN 1553-2712.
\newblock \doi{10.1111/acem.12859}.
\newblock URL \url{https://onlinelibrary.wiley.com/doi/abs/10.1111/acem.12859}.

\bibitem[Yan et~al.(2016)Yan, Ip, Raja, Gupta, Kosowsky, and
  Khorasani]{yan_yield_2016}
Zihao Yan, Ivan~K. Ip, Ali~S. Raja, Anurag Gupta, Joshua~M. Kosowsky, and Ramin
  Khorasani.
\newblock Yield of {CT} {Pulmonary} {Angiography} in the {Emergency}
  {Department} {When} {Providers} {Override} {Evidence}-based {Clinical}
  {Decision} {Support}.
\newblock \emph{Radiology}, 282\penalty0 (3):\penalty0 717--725, September
  2016.
\newblock ISSN 0033-8419.
\newblock \doi{10.1148/radiol.2016151985}.
\newblock URL \url{https://pubs.rsna.org/doi/full/10.1148/radiol.2016151985}.

\bibitem[Yang et~al.(2018)Yang, Li, Yang, and Zhou]{yang_towards_2018}
Hongmei Yang, Lin Li, Ridong Yang, and Yi~Zhou.
\newblock Towards {Automated} {Knowledge} {Discovery} of {Hepatocellular}
  {Carcinoma}: {Extract} {Patient} {Information} from {Chinese} {Clinical}
  {Reports}.
\newblock In \emph{Proceedings of the 2nd {International} {Conference} on
  {Medical} and {Health} {Informatics}}, {ICMHI} '18, pages 111--116, New York,
  NY, USA, June 2018. ACM.
\newblock ISBN 978-1-4503-6389-1.
\newblock \doi{10.1145/3239438.3239445}.
\newblock URL \url{https://doi.org/10.1145/3239438.3239445}.

\bibitem[Yasaka and Abe(2018)]{yasaka_deep_2018}
Koichiro Yasaka and Osamu Abe.
\newblock Deep learning and artificial intelligence in radiology: {Current}
  applications and future directions.
\newblock \emph{PLOS Medicine}, 15\penalty0 (11):\penalty0 e1002707, November
  2018.
\newblock ISSN 1549-1676.
\newblock \doi{10.1371/journal.pmed.1002707}.
\newblock URL
  \url{https://journals.plos.org/plosmedicine/article?id=10.1371/journal.pmed.1002707}.

\bibitem[Yim et~al.(2016{\natexlab{a}})Yim, Denman, Kwan, and
  Yetisgen]{yim_tumor_2016}
Wen-wai Yim, Tyler Denman, Sharon~W. Kwan, and Meliha Yetisgen.
\newblock Tumor information extraction in radiology reports for hepatocellular
  carcinoma patients.
\newblock \emph{AMIA Summits on Translational Science Proceedings},
  2016:\penalty0 455--464, July 2016{\natexlab{a}}.
\newblock ISSN 2153-4063.
\newblock URL \url{https://www.ncbi.nlm.nih.gov/pmc/articles/PMC5001784/}.

\bibitem[Yim et~al.(2016{\natexlab{b}})Yim, Kwan, and
  Yetisgen]{yim_tumor_2016-1}
Wen-wai Yim, Sharon~W. Kwan, and Meliha Yetisgen.
\newblock Tumor reference resolution and characteristic extraction in radiology
  reports for liver cancer stage prediction.
\newblock \emph{Journal of Biomedical Informatics}, 64:\penalty0 179--191,
  December 2016{\natexlab{b}}.
\newblock ISSN 1532-0464.
\newblock \doi{10.1016/j.jbi.2016.10.005}.
\newblock URL
  \url{http://www.sciencedirect.com/science/article/pii/S1532046416301381}.

\bibitem[Yim et~al.(2017)Yim, Kwan, and Yetisgen]{yim_classifying_2017}
Wen-wai Yim, Sharon~W. Kwan, and Meliha Yetisgen.
\newblock Classifying tumor event attributes in radiology reports.
\newblock \emph{Journal of the Association for Information Science and
  Technology}, 68\penalty0 (11):\penalty0 2662--2674, 2017.
\newblock ISSN 2330-1643.
\newblock \doi{10.1002/asi.23937}.
\newblock URL
  \url{https://asistdl.onlinelibrary.wiley.com/doi/abs/10.1002/asi.23937}.

\bibitem[Yim et~al.(2018)Yim, Kwan, Johnson, and
  Yetisgen]{yim_classification_2018}
Wen-wai Yim, Sharon~W Kwan, Guy Johnson, and Meliha Yetisgen.
\newblock Classification of hepatocellular carcinoma stages from free-text
  clinical and radiology reports.
\newblock \emph{AMIA Annual Symposium Proceedings}, 2017:\penalty0 1858--1867,
  April 2018.
\newblock ISSN 1942-597X.
\newblock URL \url{https://www.ncbi.nlm.nih.gov/pmc/articles/PMC5977638/}.

\bibitem[Young et~al.(2018)Young, Hazarika, Poria, and
  Cambria]{young_recent_2018}
T.~Young, D.~Hazarika, S.~Poria, and E.~Cambria.
\newblock Recent {Trends} in {Deep} {Learning} {Based} {Natural} {Language}
  {Processing} [{Review} {Article}].
\newblock \emph{IEEE Computational Intelligence Magazine}, 13\penalty0
  (3):\penalty0 55--75, August 2018.
\newblock ISSN 1556-6048.
\newblock \doi{10.1109/MCI.2018.2840738}.

\bibitem[Zech et~al.(2018)Zech, Pain, Titano, Badgeley, Schefflein, Su, Costa,
  Bederson, Lehar, and Oermann]{zech_natural_2018}
John Zech, Margaret Pain, Joseph Titano, Marcus Badgeley, Javin Schefflein,
  Andres Su, Anthony Costa, Joshua Bederson, Joseph Lehar, and Eric~Karl
  Oermann.
\newblock Natural {Language}–based {Machine} {Learning} {Models} for the
  {Annotation} of {Clinical} {Radiology} {Reports}.
\newblock \emph{Radiology}, 287\penalty0 (2):\penalty0 570--580, January 2018.
\newblock ISSN 0033-8419.
\newblock \doi{10.1148/radiol.2018171093}.
\newblock URL \url{https://pubs.rsna.org/doi/full/10.1148/radiol.2018171093}.

\bibitem[Zech et~al.(2019)Zech, Forde, Titano, Kaji, Costa, and
  Oermann]{zech_detecting_2019}
John Zech, Jessica Forde, Joseph~J. Titano, Deepak Kaji, Anthony Costa, and
  Eric~Karl Oermann.
\newblock Detecting insertion, substitution, and deletion errors in radiology
  reports using neural sequence-to-sequence models.
\newblock \emph{Annals of Translational Medicine}, 7\penalty0 (11), June 2019.
\newblock ISSN 2305-5839.
\newblock \doi{10.21037/atm.2018.08.11}.
\newblock URL \url{https://www.ncbi.nlm.nih.gov/pmc/articles/PMC6603352/}.

\bibitem[Zhang et~al.(2018)Zhang, Lam, Liu, Pang, Chan, and
  Tang]{zhang_development_2018}
A.~Y. Zhang, S.~S.~W. Lam, N.~Liu, Y.~Pang, L.~L. Chan, and P.~H. Tang.
\newblock Development of a {Radiology} {Decision} {Support} {System} for the
  {Classification} of {MRI} {Brain} {Scans}.
\newblock In \emph{2018 {IEEE}/{ACM} 5th {International} {Conference} on {Big}
  {Data} {Computing} {Applications} and {Technologies} ({BDCAT})}, pages
  107--115, December 2018.
\newblock \doi{10.1109/BDCAT.2018.00021}.

\bibitem[Zhang et~al.(2019)Zhang, Merck, Tsai, Manning, and
  Langlotz]{zhang_optimizing_2019}
Yuhao Zhang, Derek Merck, Emily~Bao Tsai, Christopher~D. Manning, and Curtis~P.
  Langlotz.
\newblock Optimizing the {Factual} {Correctness} of a {Summary}: {A} {Study} of
  {Summarizing} {Radiology} {Reports}.
\newblock \emph{arXiv:1911.02541 [cs.CL]}, 2019.
\newblock URL \url{http://arxiv.org/abs/1911.02541}.

\bibitem[Zhao et~al.(2018)Zhao, Fesharaki, Liu, and Luo]{zhao_using_2018}
Yiqing Zhao, Nooshin~J. Fesharaki, Hongfang Liu, and Jake Luo.
\newblock Using data-driven sublanguage pattern mining to induce knowledge
  models: application in medical image reports knowledge representation.
\newblock \emph{BMC Medical Informatics and Decision Making}, 18\penalty0
  (1):\penalty0 61, July 2018.
\newblock ISSN 1472-6947.
\newblock \doi{10.1186/s12911-018-0645-3}.
\newblock URL \url{https://doi.org/10.1186/s12911-018-0645-3}.

\bibitem[Zhu et~al.(2019)Zhu, Paschalidis, Hall, and
  Tahmasebi]{zhu_context-driven_2019}
Henghui Zhu, Ioannis~Ch. Paschalidis, Christopher Hall, and Amir Tahmasebi.
\newblock Context-{Driven} {Concept} {Annotation} in {Radiology} {Reports}:
  {Anatomical} {Phrase} {Labeling}.
\newblock \emph{AMIA Summits on Translational Science Proceedings},
  2019:\penalty0 232--241, May 2019.
\newblock ISSN 2153-4063.
\newblock URL \url{https://www.ncbi.nlm.nih.gov/pmc/articles/PMC6568085/}.

\end{thebibliography}

%\bibliographystyle{bmc-mathphys} % Style BST file (bmc-mathphys, vancouver, spbasic).
%\bibliography{bmc_article}      % Bibliography file (usually '*.bib' )

% or include bibliography directly:
% \begin{thebibliography}
% \bibitem{b1}
% \end{thebibliography}

\end{document}